\documentclass{article}
\usepackage{ijcai26}

\usepackage{times}
\usepackage{soul}
\usepackage{url}
\usepackage[hidelinks]{hyperref}
\usepackage[utf8]{inputenc}
\usepackage[small]{caption}
\usepackage{graphicx}
\usepackage{amsmath}
\usepackage{amsthm}
\usepackage{booktabs}
\usepackage{algorithm}
\usepackage{algorithmic}
\usepackage[switch]{lineno}
\usepackage[algo2e,ruled,linesnumbered]{algorithm2e}

\usepackage{multirow}
\usepackage{enumitem}
\usepackage{float}
\usepackage{subcaption}
\usepackage{caption}
\usepackage{amsfonts}
\usepackage{bm}
\usepackage[T1]{fontenc}
\usepackage{hyperref}

\usepackage{tabularx}
\usepackage{colortbl} 
\usepackage{xcolor}
\usepackage{wrapfig}
\usepackage{multicol}
\usepackage{subcaption}
\usepackage{enumitem}

\def\method{\texttt{scGTN}}

\usepackage{bm}
\usepackage{pifont}
\newcommand{\circlefir}{\ding{182}}%
\newcommand{\circlesec}{\ding{183}}%
\newcommand{\circlethi}{\ding{184}}%
\title{\method{}: Deep Siamese Graph Transformer Network for Single-cell RNA Sequencing Clustering}

\author{
Jinke Wu$^{1}$,
Yifan Wang$^{2}$,
Siyu Yi$^{1\dagger}$,
Caiyang Yu$^{1}$,
Ziyue Qiao$^{3}$,
Nan Yin$^{4}$,\\
Jiancheng Lv$^{1}$, {\normalfont and}
Wei Ju$^{1\dagger}$
\\
{\normalfont
$^{1}$Sichuan University\\
$^{2}$University of International Business and Economics\\
$^{3}$Great Bay University\\
$^{4}$The Education University of Hong Kong\\
w\_rmsl@stu.scu.edu.cn,
\{siyuyi, lvjiancheng, juwei\}@scu.edu.cn,
yifanwang@uibe.edu.cn,\\
\{yucy324, ziyuejoe, yinnan8911\}@gmail.com
}
}
\begin{document}

\maketitle
\renewcommand{\thefootnote}{\fnsymbol{footnote}}
\footnotetext[2]{Corresponding authors.}
\renewcommand{\thefootnote}{\arabic{footnote}}
\setcounter{footnote}{0}
\begin{abstract}
Single-cell RNA sequencing (scRNA-seq) serves a pivotal role in characterizing gene expression at the cellular level, enabling the identification of cell types and advancing the understanding of cellular heterogeneity. Despite the significant progress in scRNA-seq data clustering, we argue that current methods always ignore the sparsity and noise, as well as the complex intercellular structural information inherent in scRNA-seq data. Toward this end, in this paper, we propose a novel \underline{s}ingle-\underline{c}ell RNA-seq clustering framework via deep Siamese \underline{G}raph \underline{T}ransformer \underline{N}etwork (termed \method{}), which explicitly integrates gene expression profile and intercellular structural dependencies for cell clustering. In particular, we formulate scRNA-seq data as a graph and construct two augmented graph views that serve as dual views to capture complementary intercellular information. Then, a Siamese graph transformer network is employed to explicitly incorporate shortest-path information and node-wise distances for capturing richer structural relationships between cells. Finally, we employ an optimal transport strategy to guide the cell clustering in a self-supervised manner. Extensive experiments on multiple benchmark scRNA-seq datasets demonstrate that our \method{} consistently outperforms existing methods. Our code is available at \href{https://github.com/W-RMSL/scGTN}{https://github.com/W-RMSL/scGTN}.
\end{abstract}
\section{Introduction}
\label{sec::intro}
Single-cell RNA sequencing (scRNA-seq) has emerged as a fundamental technology for profiling gene expression in bioinformatics, enabling high-resolution profiling of gene expression across individual cells. Based on the scRNA-seq data, cell clustering plays a central role by organizing cells into biologically meaningful groups based on their gene expression patterns, thereby revealing the intrinsic heterogeneity of cell populations and facilitating biological interpretation of their functions and interactions~\cite{kiselev2019challenges,peng2020single}. Accordingly, clustering scRNA-seq data is particularly valuable for a range of downstream tasks, including marker gene annotation~\cite{dai2022accurate} and cell type identification~\cite{hu2020iterative}.

For effectively scRNA-seq cell clustering, numerous methods have been proposed to achieve this goal. Early approaches primarily relied on classical clustering algorithms, such as $k$-means~\cite{hicks2021mbkmeans}, hierarchical clustering~\cite{johnson1967hierarchical}, and density-based methods~\cite{petegrosso2020machine}, which are computationally efficient but are inherently limited in modeling the complex nonlinear characteristics of scRNA-seq data. With the rapid advances in deep learning, recent studies have shifted toward learning-based clustering frameworks that leverage graph neural networks (GNNs)~\cite{ju2025survey,fan2026cmgl} to extract informative representations from scRNA-seq data. By representing cells as nodes and their interactions as edges in a graph, GNN-based approaches jointly exploit gene expression features and structure characteristics to learn context-aware cell embeddings, enabling a more comprehensive cell clustering~\cite{wang2021scgnn,gan2022deep,xu2024sccdcg,xu2025scsiameseclu,zhang2026evidence}.

\begin{figure}[t]
    \centering
    \includegraphics[width=0.9\linewidth]{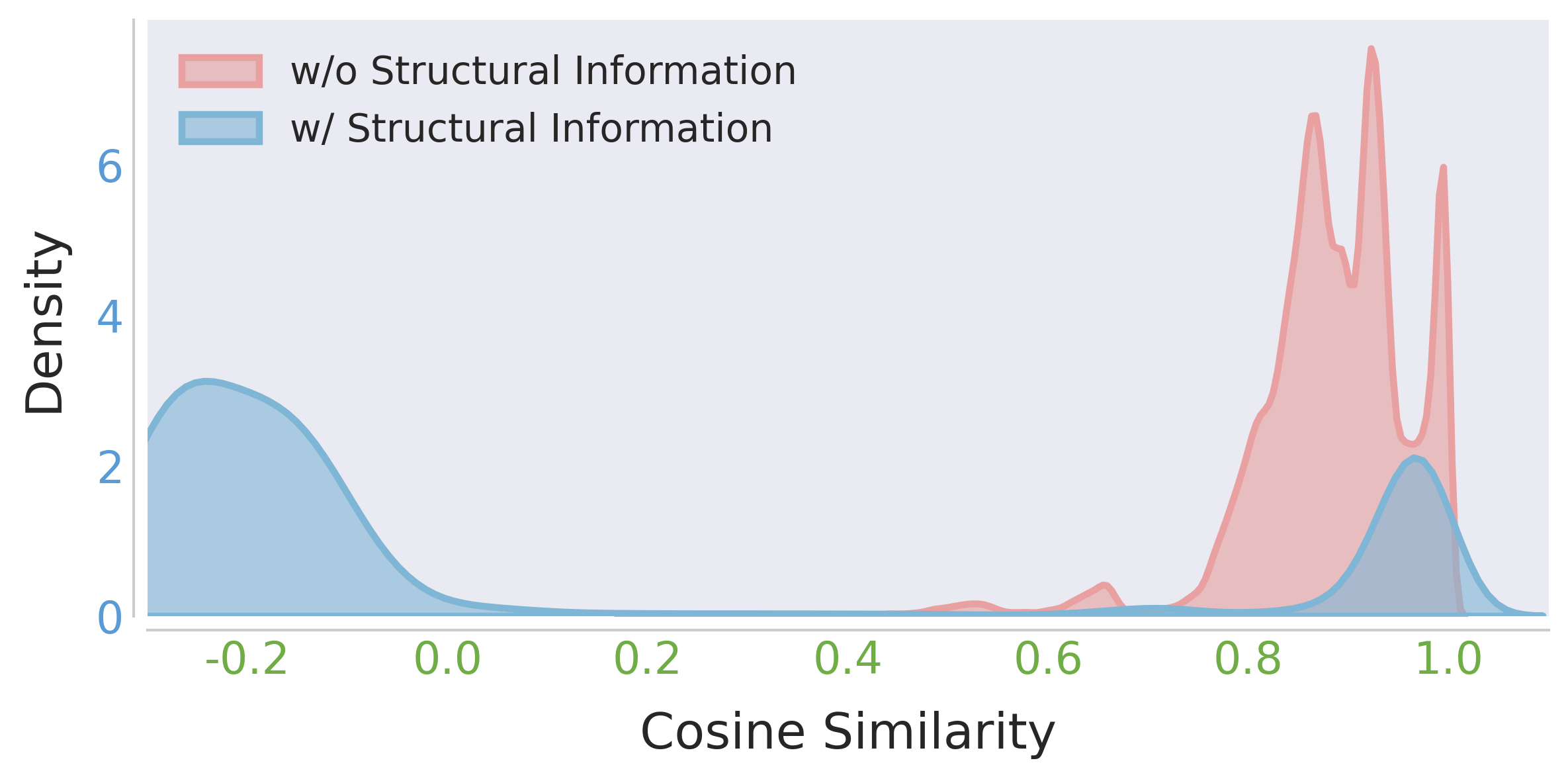} 
    \caption{Similarity distributions of cell embeddings obtained by \method{} with and without structural information on the scHuman pancreas cells dataset.}
    \label{fig:similarity_distribution}
\end{figure}

Despite the effectiveness of these methods, we argue that two major challenges remain insufficiently addressed: \circlefir{} \textbf{\textit{Ignoring the Sparsity and Noise inherent within the scRNA-seq Data.}} Arising from extensive dropout effects and measurement noise, scRNA-seq data present a high-dimensional expression matrix~\cite {ou2022matrix,wu2022network}, which substantially complicates the reliable construction of cell relationships and frequently leads to graphs that are highly sparse and noisy. \circlesec{} \textbf{\textit{Limited Exploitation of Complex Intercellular Structural Information.}} While existing GNN-based approaches primarily focus on local neighborhood aggregation, they often overlook richer structural cues embedded within the cell graph. Features such as shortest paths and node distances are rarely leveraged, resulting in an insufficient utilization of the graph’s inherent information. As shown in Figure~\ref{fig:similarity_distribution}, cell similarities derived solely from gene expression features tend to be highly homogeneous and difficult to distinguish, whereas incorporating structural information such as positional relationships and shortest-path distances enables a more discriminative characterization of intercellular relationships for cell clustering.   

Having realized the above challenges with existing methods, in this paper, we propose a novel deep Siamese \underline{\textbf{G}}raph \underline{\textbf{T}}ransformer \underline{\textbf{N}}etwork for single-cell RNA sequencing (\textbf{\method{}}), which effectively exploits complementary graph information derived from gene expression data to learn discriminative and robust cell embeddings. Specifically, given the scRNA-seq data, we formulate it into a graph and construct two augmented graph views to serve as dual branches to capture and refine the intercellular information. Then, for each branch, we introduce a graph transformer network that incorporates shortest-path information and node-wise distances to capture the complementary graph structural dependencies, enabling a more expressive modeling of complex intercellular relationships. Based on the encoded information, we utilize an optimal transport-guided self-supervised clustering method to align cluster distributions for the clustering learning process.   

To summarize, the contributions of this paper are:
\begin{itemize}[leftmargin=*] 

\item[\circlefir{}]  
\textbf{\textit{New Perspective:}} We study scRNA-seq data clustering by jointly considering gene expression profiles and cell graph structures, which effectively exploits complementary intercellular information beyond feature-level similarities. 

\item[\circlesec{}]  \textbf{\textit{Novel Methodology:}} We propose a novel framework termed \method{}, which introduces a Siamese graph transformer network to incorporate richer graph structural information, including shortest paths and node-wise distances for the cell clustering task.

\item[\circlethi{}] \textbf{\textit{Extensive Experiment:}} We conduct extensive experiments on multiple benchmark scRNA-seq datasets to evaluate the effectiveness of our \method{}. The results further show the superior clustering performance of our framework.
\end{itemize}

\section{Methodology}
\label{sec::method}
\begin{figure*}[t]
    \centering
    \includegraphics[width=\textwidth]{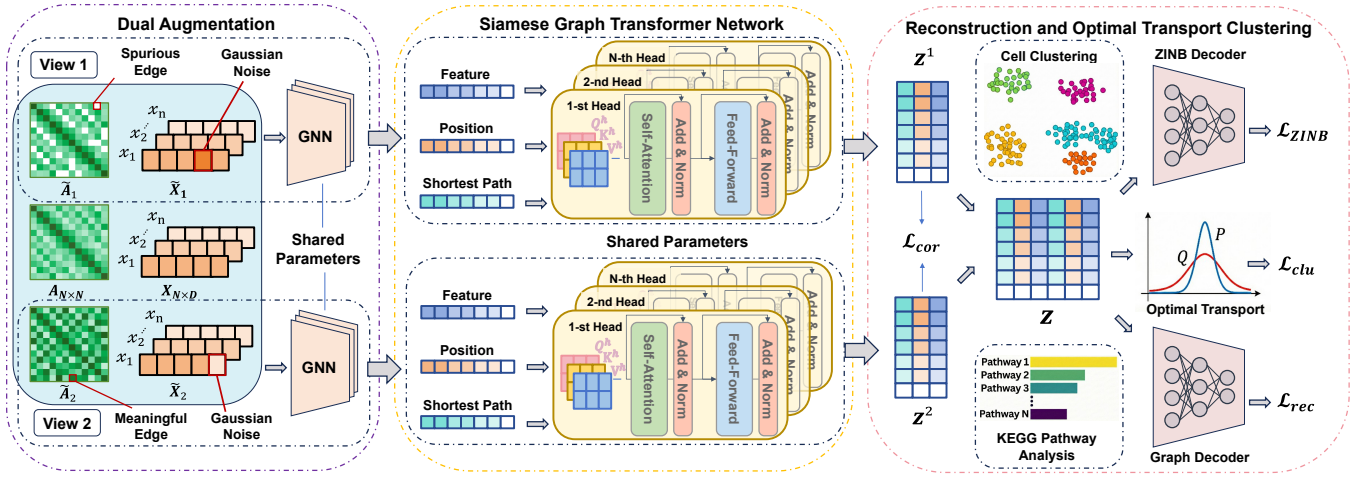}
    \caption{Illustration of the proposed \method{}, which consists of three components: (1) \textbf{Dual Augmentation for scRNA-seq Data}: Gene expression perturbation and intercellular graph structure augmentation to generate two complementary views. (2) \textbf{Siamese Graph Transformer Network Fusion}: We explicitly integrate gene expression and intercellular structure embeddings for the discriminative representation. (3) \textbf{Clustering with Optimal Transport}: We introduce a self-supervised optimal transport-guided manner to refine the cell clustering.}
    \label{fig:framework}
\end{figure*}

\subsection{Problem Definition}
Let $\bm{X}=(x_{ij})\in\mathbb{R}^{N\times D}$ denote the single-cell gene expression matrix, where each row $\bm{x}_i\in\mathbb{R}^D$ denotes the gene expression of the $i$-th cell, $N$ and $D$ denote the number of cells and gene respectively. Based on $\bm{X}$, we construct an undirected cell graph $G = (\mathcal{V}, \mathcal{E})$, where each node represents an individual cell and edges are established via a $k$-nearest neighbor ($k$NN) strategy to capture intercellular similarities quantified by the Pearson correlation coefficient~\cite{benesty2009pearson}. The adjacency matrix of the constructed graph can be $\bm{A}=(a_{ij})\in\mathbb{R}^{N\times N}$, where entry $a_{ij}=1$ if $(v_i, v_j)\in\mathcal{E}$ for cell $v_i$ and $v_j$, otherwise $a_{ij}=0$. We associate the degree matrix of the graph as $\bm{D} = \operatorname{diag}(d_1, d_2, \ldots, d_N)\in\mathbb{R}^{N\times N}$, where $d_i = \sum_{j=1}^{N} a_{ij}$ is the degree of node $v_i$. The goal of scRNA-seq clustering is to partition the cells within the graph $\mathcal{G}$ into $C$ disjoint clusters, with each node $v_i$ belonging to a unique cluster according to its gene expression features and intercellular connectivity.


\subsection{Framework Overview}
In this work, we focus on scRNA-seq data clustering and propose a Siamese graph transformer network for the task. As shown in Figure~\ref{fig:framework}, our \method{} consists of three modules: (1) \textit{Dual Augmentation Module for scRNA-seq Data}, which jointly perturbs the gene expression features and the underlying cell graph structure to generate two complementary views, thereby facilitating robust representation learning; (2) \textit{Siamese Graph Transformer Network Fusion Module}, which employs two weight-sharing Siamese encoders to process the augmented views while explicitly incorporating structural information of the cell graph; (3) \textit{Optimal Transport Clustering Module}, which formulates clustering as a distribution alignment problem and employs optimal transport to guide the learning of cluster-consistent representations.


\subsection{Dual Augmentation for scRNA-seq Data} 
To mitigate the effects of inherent sparsity and noise in scRNA-seq data, we introduce biologically reasonable augmentations by adding controlled noise to the gene expression profiles, generating two complementary views that improve the robustness of the learned representations. 

\smallskip\noindent\textbf{Gene Expression Perturbation.} Given the single-cell gene expression matrix $\bm{X}$, we employ Gaussian noise to perturb the gene expression profiles, which can simulate the natural variability of the gene expression profile as augmentation for the clustering task. The perturbation can be formulated as:
\begin{equation}
    \tilde{\bm{X}} = \{\bm{x}_1\odot\bm{m},\dots,\bm{x}_{N}\odot\bm{m}\},
\end{equation}
where $\bm{m}$ denotes the random mask drawn from a Gaussian distribution. In practice, we set the distribution as $\mathcal{N}(1,0.1)$ following the previous work~\cite{xu2025scsiameseclu}. The two augmented gene matrices can be generated as $\tilde{\bm{X}}^1$ and $\tilde{\bm{X}}^2$.

\smallskip\noindent\textbf{Intercellular Graph Structure Augmentation.} In addition to perturbing the gene expression profiles, enhancing the intercellular structure is essential for improving the model's ability to capture the relationships between cells. Specifically, given the constructed cell graph, we remove spurious edges to refine the graph structure and preserve the most biologically relevant connections between cells as one view:
\begin{equation}
    \tilde{\bm{A}}^1=\bm{D}^{-\frac{1}{2}}(\bm{A}\odot\bm{M}+\bm{I})\bm{D}^{-\frac{1}{2}},
\end{equation}
where $\bm{M}$ denotes the mask matrix based on pairwise cosine similarity. In practice, we identify and remove 10\% of edges with the lowest similarity values using this mask. For another view, we enhance the intercellular structure by considering graph diffusion, which propagates information across the graph to strengthen meaningful cell connections:
\begin{equation}
    \tilde{\bm{A}}^2=\eta(\bm{I}-(1-\eta)\bm{D}^{-\frac{1}{2}}\bm{A}\bm{D}^{-\frac{1}{2}})^{-1},
\end{equation}
where $\eta$ denotes the diffusion coefficient to control the information propagation within the cell graph.



\subsection{Siamese Graph Transformer Network Fusion}
To further utilize the richer structural information from the intercellular graph, we introduce a Siamese graph transformer network~\cite{wang2025deep} for the two augmented views and adaptively fuse the encoded feature to learn discriminative cell representations.

\smallskip\noindent\textbf{Gene Expression Embedding.}
Since cells with similar gene expression profiles are assumed to be connected in the graph, we leverage the gene expression profiles of connected cells to enhance the learning of discriminative gene expression embeddings. In particular, we employ two Siamese graph encoders to produce the cell embedding as:
\begin{equation}
     \bm{Z}^*=\mathcal{G}(\tilde{A}^*,\tilde{X}^*), *\in\{1,2\},
\end{equation}
where $\mathcal{G}(\cdot,\cdot)$ denotes the Siamese graph encoder, which is shared across the two augmented views. The resulting encoded gene expression matrix $\bm{Z}^*\in\mathbb{R}^{N\times d}$ represents the learned feature embeddings for each cell, where $d$ indicates the dimensionality of the feature space. Then, for each cell node with feature $\bm{z}_i^*$, we select the top-$t$ neighbors with the most similar features and concatenate them into a sequence $\bm{S}_i^*\in\mathbb{R}^{(t+1)\times D}$. The sequence is then encoded as:
\begin{equation}
    \bm{E}_i^*=\mathcal{F}_{ge}(\bm{S}_i^*)\in\mathbb{R}^{(t+1)\times d},
\end{equation}
where $\mathcal{F}_{ge}(\cdot)$ denotes the gene expression encoding function.


\smallskip\noindent\textbf{Explicit Intercellular Structure Embedding.}
To explicitly capture the intercellular structure within scRNA-seq data, we incorporate both the position embedding and shortest path embedding of the central node and its neighbors. Specifically, we define the relative positions of the central node as $\operatorname{Pos}(\bm{z}_i^*)=0$ and assign position values to its neighbors based on their intimacy to the central node, with nodes that are more similar to 
the central node receiving position values closer to 0. The position embedding is formulated as:
\begin{equation}
    \bm{P}_i^*=\mathcal{F}_{pos}(\operatorname{Pos}(\bm{z}_i^*))\in\mathbb{R}^{(t+1)\times d},
\end{equation}
where $\mathcal{F}_{pos}(\cdot)$ denotes the position mapping function. In addition to the intimacy between two nodes, the shortest path characterizes the connection structure between them and explicitly the crucial intercellular structure. Specifically, the shortest path between two nodes within the constructed cell graph can be defined as $\operatorname{Sp}(\bm{z}_i^*,\bm{z}_j^*)$. Based on this, we formulate the shortest path embedding as:
\begin{equation}
    \bm{H}_i^*=\mathcal{F}_{sp}(\operatorname{Sp}(\bm{z}_i^*,\bm{z}_j^*))\in\mathbb{R}^{(t+1)\times d},
\end{equation}
where $\mathcal{F}_{sp}(\cdot)$ denote the shortest path mapping function.





\smallskip\noindent\textbf{Siamese Graph Transformer Refinement.}
We integrate the gene expression and intercellular structure embedding as:
\begin{equation}
    \bm{Y}_i^* = \bm{E}_i^* + \bm{P}_i^* + \bm{H}_i^*\in\mathbb{R}^{(t+1)\times d}.
\end{equation}
Then we leverage the attention mechanism to encode the integrated embedding, which can be formulated as:
\begin{equation}
    \!\!\!\!\operatorname{Attention}(\bm{Y}_i)=\operatorname{Softmax}\left(\frac{\bm{W}_Q\bm{Y}_i(\bm{W}_K\bm{Y}_i)^\top}{\sqrt{d}}\right)\bm{W}_V\bm{Y}_i
\end{equation}
where $\bm{W}_Q,\bm{W}_K,\bm{W}_V\in\mathbb{R}^{d\times d}$ denote the query, key and value matrices respectively. We employ the multi-attention mechanism with feed-forward networks (FFN) and residual connections as a transformer layer. Through stacking $L$-layers, the output feature can be $\bm{Y}_i^{L,*}\in\mathbb{R}^{(t+1)\times d}$. The final cell node embedding can be refined as:
\begin{equation}
    \hat{\bm{z}}_i^*=\frac{1}{t+1}\sum\nolimits_{t=0}^t\bm{Y}_i^{L,*}[t,:]\in\mathbb{R}^d.
\end{equation}
We further introduce the correlation loss to prevent feature collapse and redundancy between two views, defined as:
\begin{equation}
\begin{aligned}
    \bm{R}_{ij}&=\frac{\hat{\bm{z}}_i^1(\hat{\bm{z}}_j^2)^\top}{\|\hat{\bm{z}}_i^1\|\|\hat{\bm{z}}_j^2)\|},\\
    \mathcal{L}_{\text{cor}}=\sum\nolimits_{i}(1&-\bm{R}_{ii})^2+\sum\nolimits_i\sum\nolimits_{j\neq i}\bm{R}_{ij}^2.
\end{aligned}
\end{equation}
The final cell embedding $\hat{\bm{Z}}$ can be fused from two views as: 
\begin{equation}
    \hat{\bm{Z}}=\frac{1}{2}(\hat{\bm{Z}}^1+\hat{\bm{Z}}^2).
\end{equation}


\subsection{Clustering with Optimal Transport} Given the refined cell embedding, we adopt a self-supervised strategy for the scRNA-seq data clustering. Specifically, we measure the similarity between cell embedding $\hat{\bm{z}}_i$ and the clustering center $\bm{\mu}_c$ as the Student's $t$-distribution~\cite{maaten2008visualizing}, formulated as:
\begin{equation}
    q_{ic}=\frac{(1+\|\hat{\bm{z}}_i-\bm{\mu}_c\|/\theta)^{-\frac{1+\theta}{2}}}{\sum_{c'}(1+\|\hat{\bm{z}}_i-\bm{\mu}_{c'}\|/\theta)^{-\frac{1+\theta}{2}}},
\end{equation}
where $\bm{Q}=[q_{ic}]\in\mathbb{R}_{+}^{N\times C}$ denotes the cluster assignment of the cell and $\theta$ is the degrees of freedom in the Student’s $t$-distribution. Following the previous work~\cite{bo2020structural}, we sharpen the cluster assignment as the target distribution $\bm{P}=[p_{ic}]\in\mathbb{R}_{+}^{N\times C}$, which can be calculated as:
\begin{equation}
    p_{ic}=\frac{q_{ic}^2/f_c}{\sum q_{ic'}^2/f_{c'}},
\end{equation}
where $f_c=\sum_iq_{ic}$ denotes the soft cluster frequency. We further introduce the optimal transport strategy to align the cluster distribution with the mixing proportions, which ensures the robust clustering and prevents the degenerate solution that all cells allocate to a single label:
\begin{equation}
\begin{aligned}
    &\min_{\bm{P}}-\bm{P}*(\log\bm{Q})\\
    &\operatorname{s.t.} \bm{P}\bm{1}_C = \bm{1}_N \ \text{ and } \ P^\top\bm{1}_N = N\bm{\pi},
\end{aligned}
\end{equation}
where $\bm{P}$ and $-\log\bm{Q}$ denote the transport plan and cost matrix, respectively. Here $\bm{\pi}$ in the constraints represents the proportion of cells assigned to each cluster, which can be derived from the intermediate clustering results. In practice, we utilize the Sinkhorn distance method~\cite{sinkhorn1967diagonal} for the optimization by introducing entropy constraint $H(\cdot)$ with a Lagrange multiplier $\lambda$ as:
\begin{equation}
\begin{aligned}
    &\min_{\bm{P}}-\bm{P}*(\log\bm{Q})-\frac{1}{\lambda}H(\bm{P})\\
    &\operatorname{s.t.} \bm{P}\bm{1}_C = \bm{1}_N \ \text{ and } \ P^\top\bm{1}_N = N\bm{\pi}.
\end{aligned}
\end{equation}
Therefore, we can iteratively optimize the $\bm{P}$ as:
\begin{equation}
    \hat{\bm{P}}^{(t)}=\operatorname{diag}(\bm{u}^{(t)}) \bm{Q}^\lambda \operatorname{diag}(\bm{v}^{(t)}),
\end{equation}
where we iteratively update  $\bm{u}^{(t)}$ and $\bm{v}$ as $\bm{1}_N / (\bm{Q}^{\lambda}\bm{v}^{(t-1)})$ and $N_\pi / (\bm{Q}^{\lambda}\bm{u}^{(t)})$ at iteration $t$ with $N_\pi$ representing the the total mass of the target distribution and the initial value setting as $\bm{v}^{(0)}=\bm{1}_N$. During the training process, we fixed $\hat{\bm{P}}$ and align $\bm{Q}$ with the $\hat{\bm{P}}$ as the clustering loss:
\begin{equation}
    \mathcal{L}_{\text{clu}}=\operatorname{KL}(\hat{\bm{P}}\|\bm{Q})=\sum\nolimits_{i=1}^N\sum\nolimits_{c=1}^C\hat{p}_{ic}\log\frac{\hat{p}_{ic}}{q_{ic}},
\end{equation}


\subsection{Overall Optimization}
To preserve the intercellular structure information for cell clustering, we reconstruct the cell graph with the learned cell embedding and the Mean Squared Error (MSE) loss can be:
\begin{equation}
    \mathcal{L}_{\text{rec}}=\|\bm{A}-\bm{M}\|_F^2,\ \bm{M}_{ij}=\frac{1}{2}\left(\frac{\hat{\bm{z}}_i\cdot\hat{\bm{z}}_j}{\|\hat{\bm{z}}_i\|\|\hat{\bm{z}}_j\|}+1\right).
\end{equation}
Since scRNA-seq data exhibits sparsity and overdispersion, we further incorporate the Zero-Inflated Negative Binomial (ZINB) loss~\cite{eraslan2019single,xu2025scsiameseclu} to model the excess zeros and variability. Finally, the overall objective of our \method{} can be defined as:
\begin{equation}
    \mathcal{L}=\mathcal{L}_{\text{clu}}+\alpha\mathcal{L}_{\text{cor}}+\beta\mathcal{L}_{\text{rec}}+\gamma\mathcal{L}_{\text{ZINB}},
\end{equation}
where $\alpha,\beta$ and $\gamma$ denote the trade-off parameter to balance the contribution of each component.

\section{Experiments}
\label{sec::experiment}
\begin{table*}[t]
\centering

\newcommand{\res}[2]{%
  #1{\scriptsize\,\ensuremath{\pm}\,#2}%
}
\newcommand{\best}[2]{%
  \textbf{#1{\scriptsize\,\ensuremath{\pm}\,#2}}%
}
\newcommand{\second}[2]{%
  \underline{#1{\scriptsize\,\ensuremath{\pm}\,#2}}%
}

\newcommand{\head}[1]{\mbox{\textbf{#1}}}
\newcolumntype{Y}{>{\centering\arraybackslash}X}

\renewcommand{\arraystretch}{0.95} 
\setlength{\tabcolsep}{1pt}

\resizebox{\linewidth}{!}{
\begin{tabularx}{1.5\linewidth}{c c | YYYYYYYYYYY} 
\toprule

\textbf{Datasets} & \textbf{Metric}
& \head{pcaReduce} & \head{DEC} & \head{contrastive-sc} & \head{scNAME} & \head{scDeepCluster}
& \head{scDSC} & \head{AttentionAE-sc} & \head{scGNN} & \head{scCDCG} & \head{scSiameseClu} & \head{Ours} \\
\midrule

\multirow{3}{*}{\shortstack{\textbf{Muraro Human}\\\textbf{Pancreas cells}}}
& ACC & \res{50.56}{3.4} & \res{72.22}{5.5} & \res{81.85}{4.6} & \res{80.36}{7.9} & \res{74.70}{2.7} & \res{79.42}{1.4} & \res{93.84}{2.7} & \res{79.32}{4.0} & \res{92.65}{1.9} & \second{94.95}{0.1} & \best{96.02}{0.5} \\
& NMI & \res{54.83}{1.7} & \res{76.29}{2.7} & \res{77.73}{3.3} & \res{79.12}{2.7} & \res{79.32}{0.3} & \res{75.89}{0.6} & \second{87.80}{2.3} & \res{78.76}{5.6} & \res{86.81}{1.0} & \res{86.19}{0.3} & \best{89.15}{1.1} \\
& ARI & \res{36.61}{2.3} & \res{61.76}{5.6} & \res{75.22}{10.1} & \res{76.64}{10.4} & \res{64.59}{2.5} & \res{75.42}{1.2} & \res{90.74}{4.1} & \res{78.87}{3.2} & \res{91.37}{1.2} & \second{91.59}{0.3} & \best{93.10}{0.9} \\
\midrule

\multirow{3}{*}{\shortstack{\textbf{Human Pancreas}\\\textbf{cells 1}}}
& ACC & \res{23.63}{0.2} & \res{40.64}{1.9} & \res{69.67}{7.8} & \res{84.20}{6.0} & \res{66.39}{3.8} & \res{73.10}{2.3} & \res{81.56}{1.8} & \res{55.10}{2.1} & \res{92.15}{0.8} & \second{95.12}{0.4} & \best{97.21}{0.3} \\
& NMI & \res{42.58}{0.4} & \res{62.35}{0.6} & \res{69.21}{3.3} & \res{65.74}{4.0} & \res{69.42}{12.5} & \res{60.50}{4.2} & \res{82.62}{3.0} & \res{64.03}{0.9} & \res{86.35}{0.9} & \second{89.50}{0.6} & \best{93.23}{0.2} \\
& ARI & \res{1.98}{0.5} & \res{26.07}{1.3} & \res{52.00}{8.7} & \res{63.31}{7.4} & \res{47.92}{3.0} & \res{69.81}{1.6} & \res{71.84}{8.5} & \res{38.32}{0.8} & \res{92.83}{0.6} & \second{93.40}{0.5} & \best{96.61}{1.5} \\
\midrule

\multirow{3}{*}{\shortstack{\textbf{Human Pancreas}\\\textbf{cells 2}}}
& ACC & \res{37.56}{0.4} & \res{45.26}{1.8} & \res{60.01}{5.7} & \res{63.33}{2.0} & \res{62.88}{5.5} & \res{80.40}{4.0} & \res{80.40}{4.0} & \res{57.62}{2.3} & \res{84.66}{1.1} & \second{89.45}{0.9} & \best{91.13}{0.9} \\
& NMI & \res{50.81}{0.3} & \res{66.32}{2.1} & \res{62.25}{2.3} & \res{58.05}{1.6} & \res{73.97}{2.2} & \res{79.44}{1.7} & \res{82.16}{10.9} & \res{79.32}{2.3} & \res{83.44}{2.6} & \second{86.80}{1.2} & \best{90.63}{1.2} \\
& ARI & \res{19.23}{0.5} & \res{39.14}{1.8} & \res{47.10}{5.3} & \res{32.12}{1.5} & \res{64.56}{7.5} & \res{82.85}{6.5} & \res{73.20}{10.5} & \res{79.96}{1.7} & \res{85.30}{1.6} & \second{89.20}{1.1} & \best{92.38}{1.4} \\
\midrule

\multirow{3}{*}{\shortstack{\textbf{Human Pancreas}\\\textbf{cells 3}}}
& ACC & \res{35.92}{0.4} & \res{43.56}{1.5} & \res{57.47}{5.0} & \res{83.20}{8.7} & \res{73.65}{3.1} & \res{80.12}{10.0} & \second{90.96}{2.4} & \res{67.94}{4.6} & \res{88.04}{0.3} & \res{89.45}{0.5} & \best{95.31}{1.4} \\
& NMI & \res{51.22}{0.3} & \res{64.51}{0.8} & \res{66.84}{2.3} & \res{80.56}{5.6} & \res{75.02}{1.2} & \res{75.87}{8.4} & \second{87.00}{1.8} & \res{62.60}{1.8} & \res{82.21}{1.1} & \res{86.20}{0.6} & \best{91.37}{0.3} \\
& ARI & \res{26.40}{4.5} & \res{42.73}{1.0} & \res{47.90}{3.4} & \res{81.08}{5.7} & \res{63.31}{2.3} & \res{82.85}{8.5} & \res{89.78}{2.5} & \res{58.41}{1.9} & \res{90.78}{0.6} & \second{92.34}{0.8} & \best{95.25}{0.5} \\
\midrule

\multirow{3}{*}{\shortstack{\textbf{Mouse Pancreas}\\\textbf{cells 2}}}
& ACC & \res{28.97}{0.3} & \res{34.19}{2.3} & \res{59.55}{5.7} & \res{82.36}{8.0} & \res{69.91}{2.9} & \res{77.35}{2.3} & \res{81.92}{6.5} & \res{43.03}{3.0} & \second{93.97}{0.2} & \res{92.45}{0.4} & \best{95.11}{1.2} \\
& NMI & \res{48.93}{0.1} & \res{54.08}{0.9} & \res{62.34}{2.4} & \res{78.00}{3.9} & \res{59.58}{2.6} & \res{62.50}{0.4} & \res{81.88}{5.1} & \res{55.87}{1.9} & \second{88.02}{0.2} & \res{86.50}{0.5} & \best{88.59}{0.6} \\
& ARI & \res{12.80}{0.1} & \res{18.54}{1.1} & \res{49.75}{8.5} & \res{76.56}{11.4} & \res{60.14}{3.6} & \res{65.81}{1.4} & \res{82.14}{7.8} & \res{26.40}{2.5} & \second{92.61}{0.3} & \res{90.80}{0.6} & \best{93.43}{0.8} \\
\midrule

\multirow{3}{*}{\textbf{CITE-CMBC}}
& ACC & \res{28.19}{0.1} & \res{41.88}{5.4} & \res{52.70}{5.0} & \res{68.06}{13.3} & \res{70.80}{2.5} & \res{68.79}{0.9} & \res{61.18}{7.9} & \res{66.71}{4.0} & \res{71.45}{1.8} & \second{74.70}{0.5} & \best{79.20}{0.5} \\
& NMI & \res{28.36}{0.2} & \res{46.87}{9.2} & \res{64.83}{0.8} & \res{63.23}{13.2} & \res{72.53}{0.7} & \res{64.21}{3.3} & \res{62.02}{9.6} & \res{61.70}{3.1} & \second{74.77}{1.7} & \res{68.71}{0.7} & \best{76.50}{0.4} \\
& ARI & \res{5.10}{0.1} & \res{23.55}{10.3} & \res{47.88}{3.7} & \res{58.99}{21.3} & \res{56.23}{4.1} & \res{52.51}{2.2} & \res{44.20}{11.6} & \res{61.20}{3.1} & \res{61.46}{1.4} & \second{67.56}{1.1} & \best{69.80}{0.5} \\
\midrule

\multirow{3}{*}{\shortstack{\textbf{Human Liver}\\\textbf{cells}}}
& ACC & \res{34.92}{0.2} & \res{46.32}{5.5} & \res{79.31}{2.1} & \res{74.55}{5.3} & \res{63.44}{0.7} & \res{70.90}{2.2} & \res{79.46}{6.7} & \res{68.65}{3.2} & \res{75.34}{1.7} & \second{88.33}{1.7} & \best{94.50}{0.9} \\
& NMI & \res{32.07}{0.6} & \res{52.24}{5.0} & \res{85.11}{0.1} & \res{77.42}{12.0} & \res{74.60}{2.5} & \res{71.63}{2.8} & \res{82.86}{4.5} & \res{62.33}{2.1} & \res{79.34}{2.6} & \second{88.82}{1.2} & \best{90.15}{1.1} \\
& ARI & \res{6.34}{14.4} & \res{31.04}{4.7} & \res{89.96}{0.1} & \res{79.91}{2.9} & \res{58.58}{10.1} & \res{75.34}{3.6} & \res{79.16}{11.8} & \res{65.41}{1.3} & \res{81.26}{2.7} & \second{91.90}{0.5} & \best{94.10}{0.5} \\
\bottomrule
\end{tabularx}
}
\caption{Clustering performance across seven benchmark datasets. The results are reported as mean $\pm$ standard deviation, where the best results are highlighted in \textbf{bold} and the second-best are \underline{underlined}.}
\label{tab:main_result}
\end{table*}
\begin{figure*}[t]
    \centering
    \begin{subfigure}[b]{0.14\linewidth}
        \includegraphics[width=\linewidth]{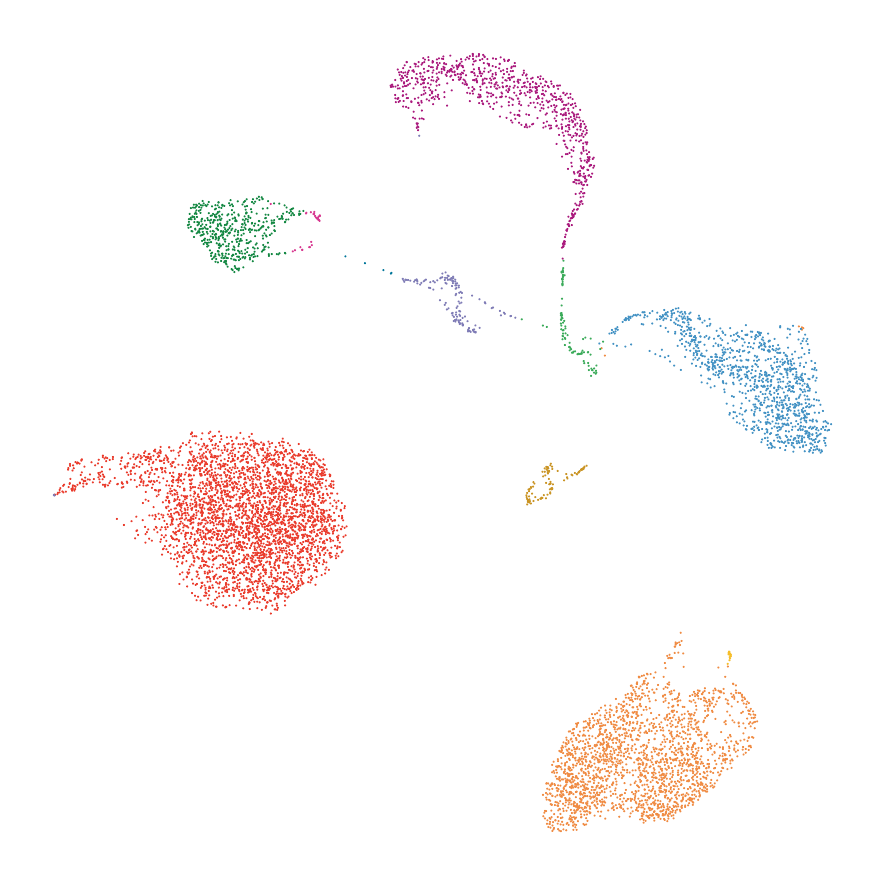}
        \caption{scDeepCluster}
    \end{subfigure}
    \hfill
    \begin{subfigure}[b]{0.16\linewidth}
        \includegraphics[width=\linewidth]{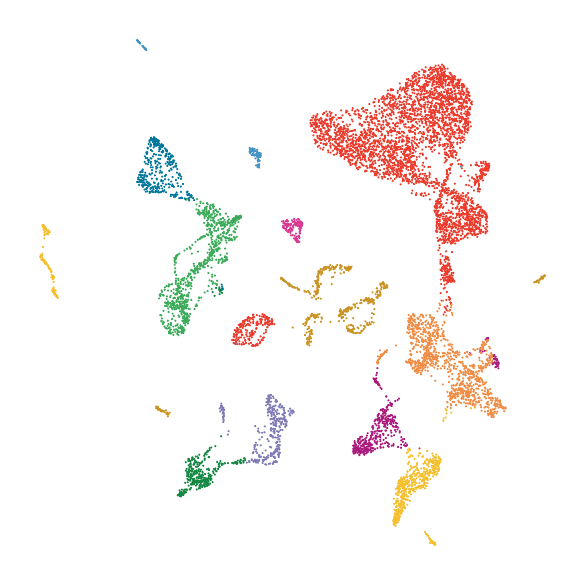}
        \caption{scDSC}
    \end{subfigure}
    \hfill
    \begin{subfigure}[b]{0.16\linewidth}
        \includegraphics[width=\linewidth]{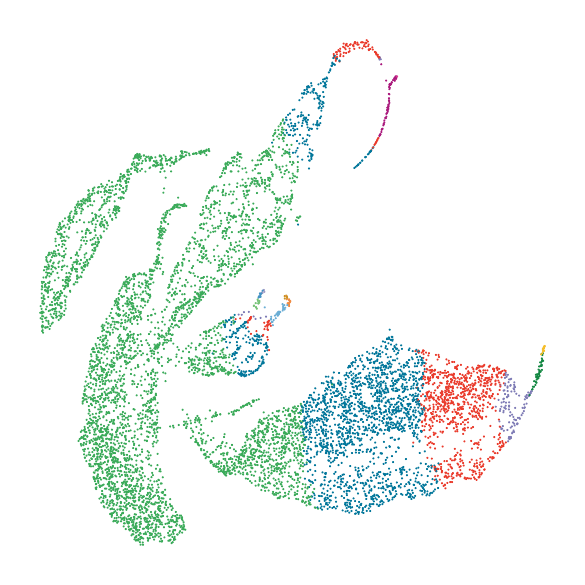}
        \caption{scGNN}
    \end{subfigure}
    \hfill
    \begin{subfigure}[b]{0.16\linewidth}
        \includegraphics[width=\linewidth]{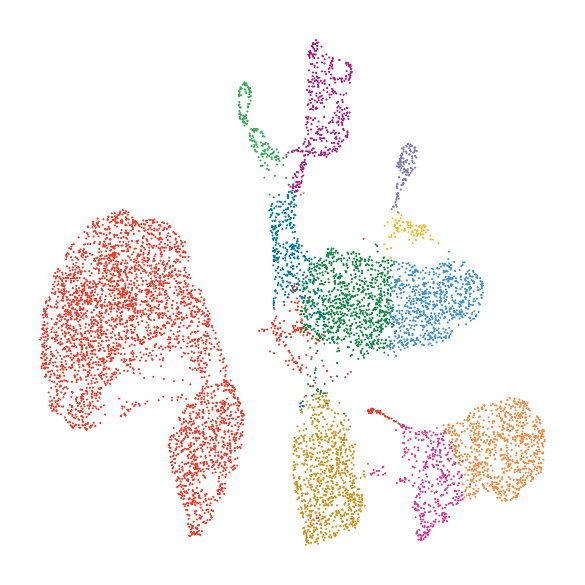}
        \caption{scCDCG}
    \end{subfigure}
    \hfill
    \begin{subfigure}[b]{0.16\linewidth}
        \includegraphics[width=\linewidth]{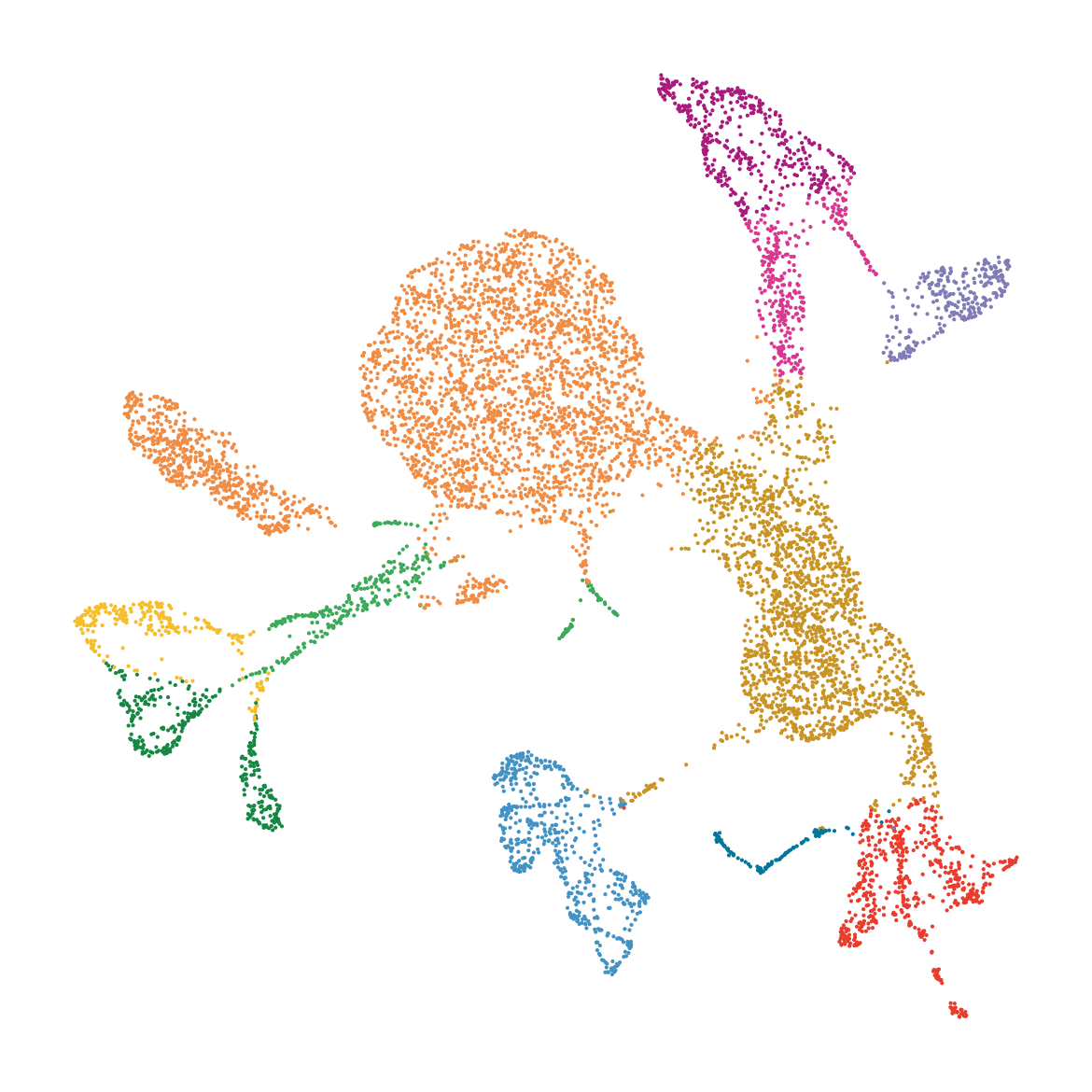}
        \caption{scSiameseClu}
    \end{subfigure}
    \hfill
    \begin{subfigure}[b]{0.16\linewidth}
        \includegraphics[width=\linewidth]{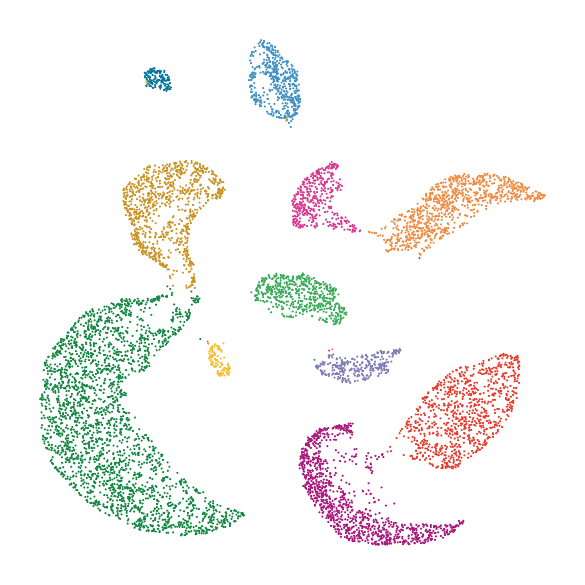}
        \caption{\method{}}
    \end{subfigure}
    \caption{UMAP visualizations of \method{} and five baseline methods on the Human Liver cells dataset. Each point represents a cell, and colors indicate the predicted cell type.}

    \label{fig:UMAP_comparison}
\end{figure*}

\subsection{Experimental Setup}

\smallskip\noindent\textbf{Benchmark Datasets}. To evaluate the robustness of our model for clustering across species and diverse biological contexts, we conduct experiments on seven public single-cell benchmark datasets, covering both human and mouse samples and including pancreas, liver, and peripheral immune cell populations. 
In terms of data scale, the number of cells ranges from 1{,}064 to 8{,}617, the number of genes ranges from 2{,}000 to 20{,}125, and the expression matrices exhibit sparsity rates of approximately 73.02\%--93.26\%. To standardize the input dimensionality and focus on more informative features, we select either 1{,}500 or 2{,}000 highly variable genes (HVGs) for each dataset as the model input. Detailed statistics are summarized in the Table~\ref{tab:dataset_summary} of Appendix~\ref{sec:setup}.

\smallskip\noindent\textbf{Baselines}. To highlight the advantages of \method{}, we compare it with ten representative clustering approaches. Specifically, pcaReduce~\cite{vzurauskiene2016pcareduce} is included as a representative traditional clustering method. DEC~\cite{xie-et-al:dec}, contrastive-sc ~\cite{ciortan2021contrastive}, scNAME~\cite{wan-et-al:scname}, scDeepCluster~\cite{tian-et-al:model-based-scrnaseq} and AttentionAE-sc~\cite{li-et-al:attention-scrnaseq} are deep learning--based clustering methods, which mainly learn representations via autoencoders or contrastive learning and then perform clustering. In addition, scGNN~\cite{wang2021scgnn}, scDSC~\cite{gan2022deep},  scCDCG~\cite{xu2024sccdcg} and scSiamese~\cite{xu2025scsiameseclu} are deep graph clustering methods that incorporate GNNs and structural information for representation learning and clustering. Detailed experimental settings and implementation details are provided in the Appendix~\ref{sec:setup}.

\smallskip\noindent\textbf{Evaluation Metrics}. To evaluate the effectiveness of the proposed method, we assess clustering performance using three widely adopted metrics: clustering accuracy (ACC), normalized mutual information (NMI)~\cite{strehl-ghosh:cluster-ensembles}, and adjusted Rand index (ARI)~\cite{vinh-et-al:ari}.

\begin{figure*}[t]
    \centering
    \begin{subfigure}[b]{0.19\linewidth}
        \includegraphics[width=\linewidth]{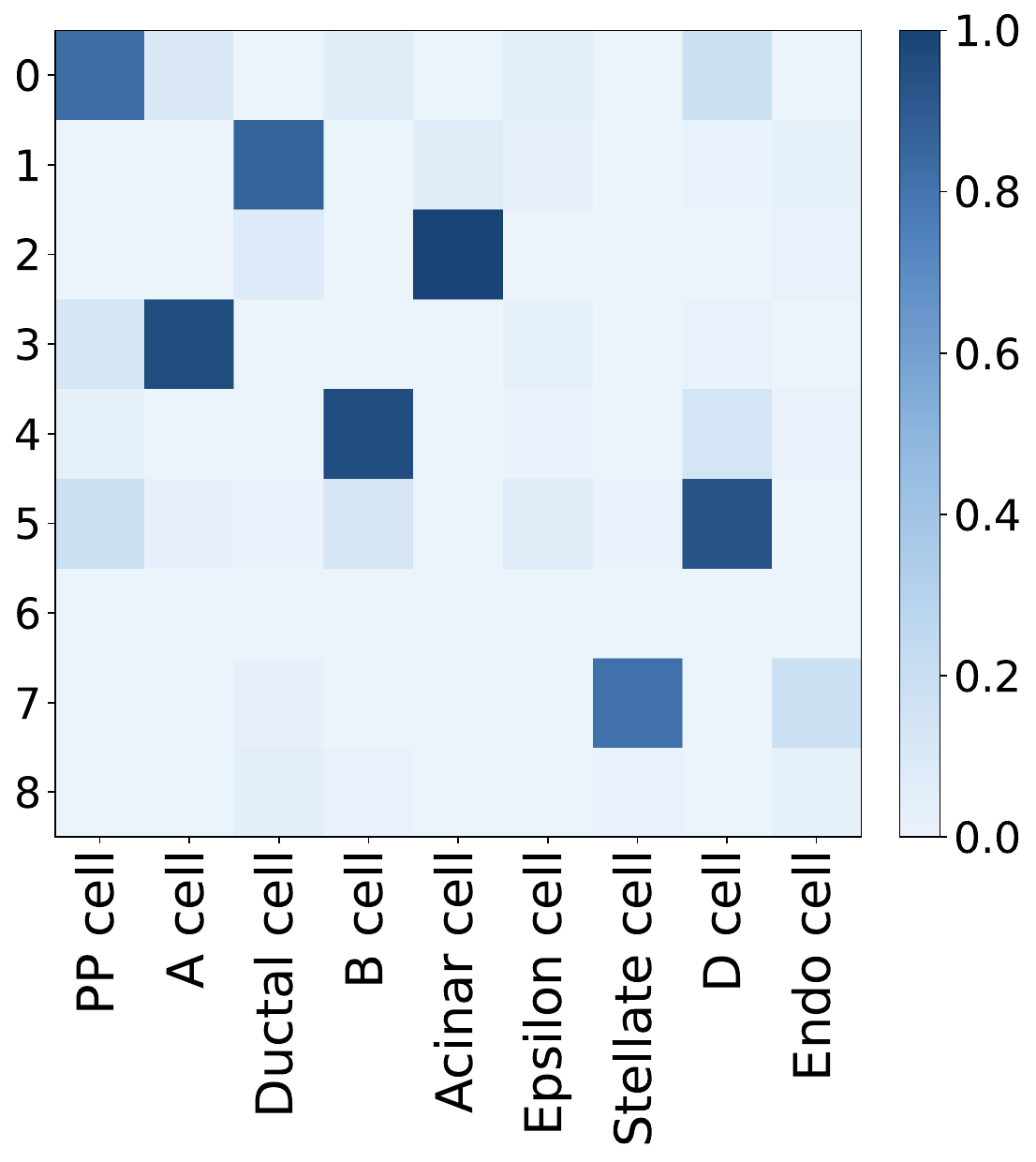}
        \caption{scDSC}
    \end{subfigure}\hfill
    \begin{subfigure}[b]{0.19\linewidth}
        \includegraphics[width=\linewidth]{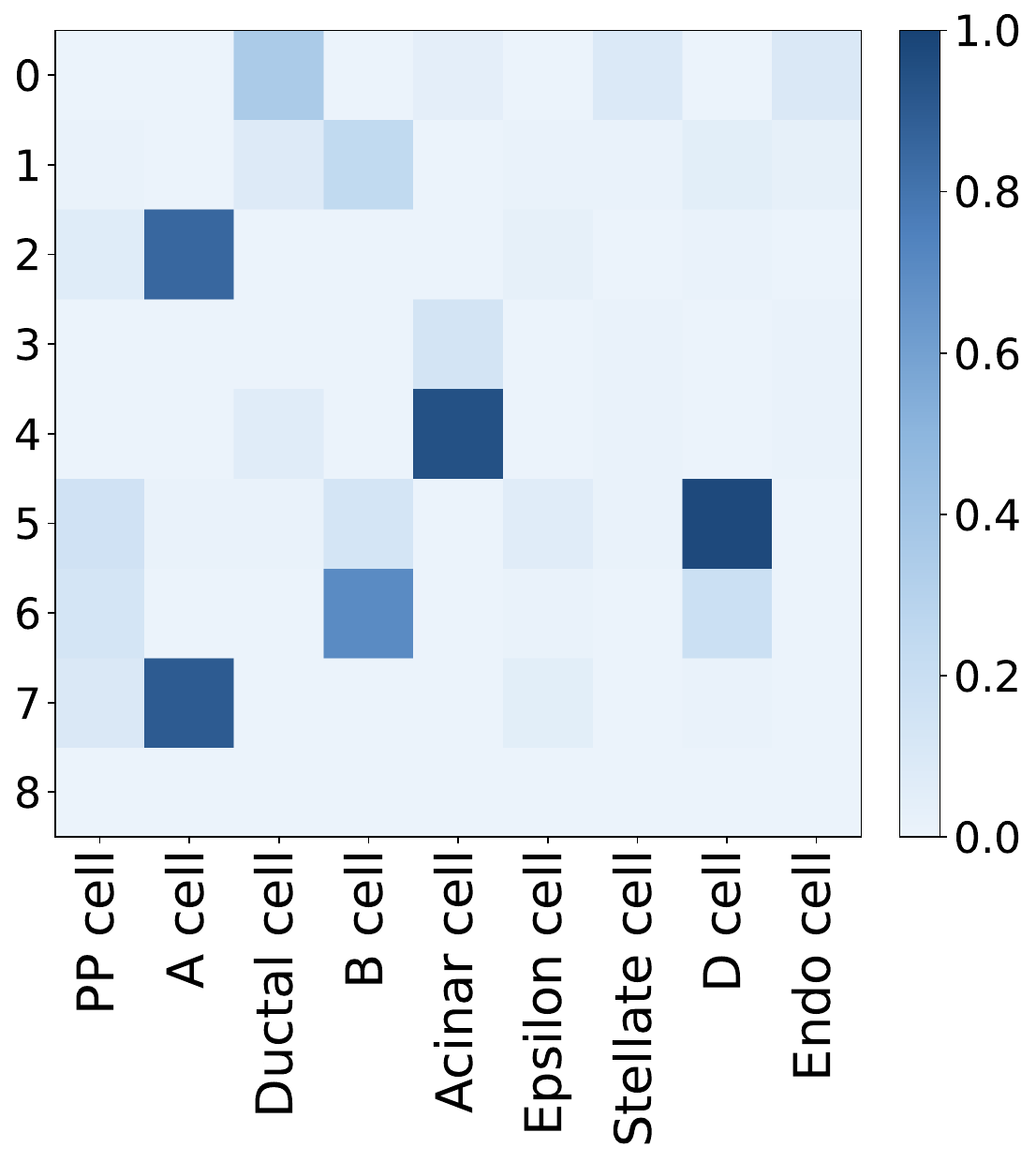}
        \caption{scGNN}
    \end{subfigure}\hfill
    \begin{subfigure}[b]{0.19\linewidth}
        \includegraphics[width=\linewidth]{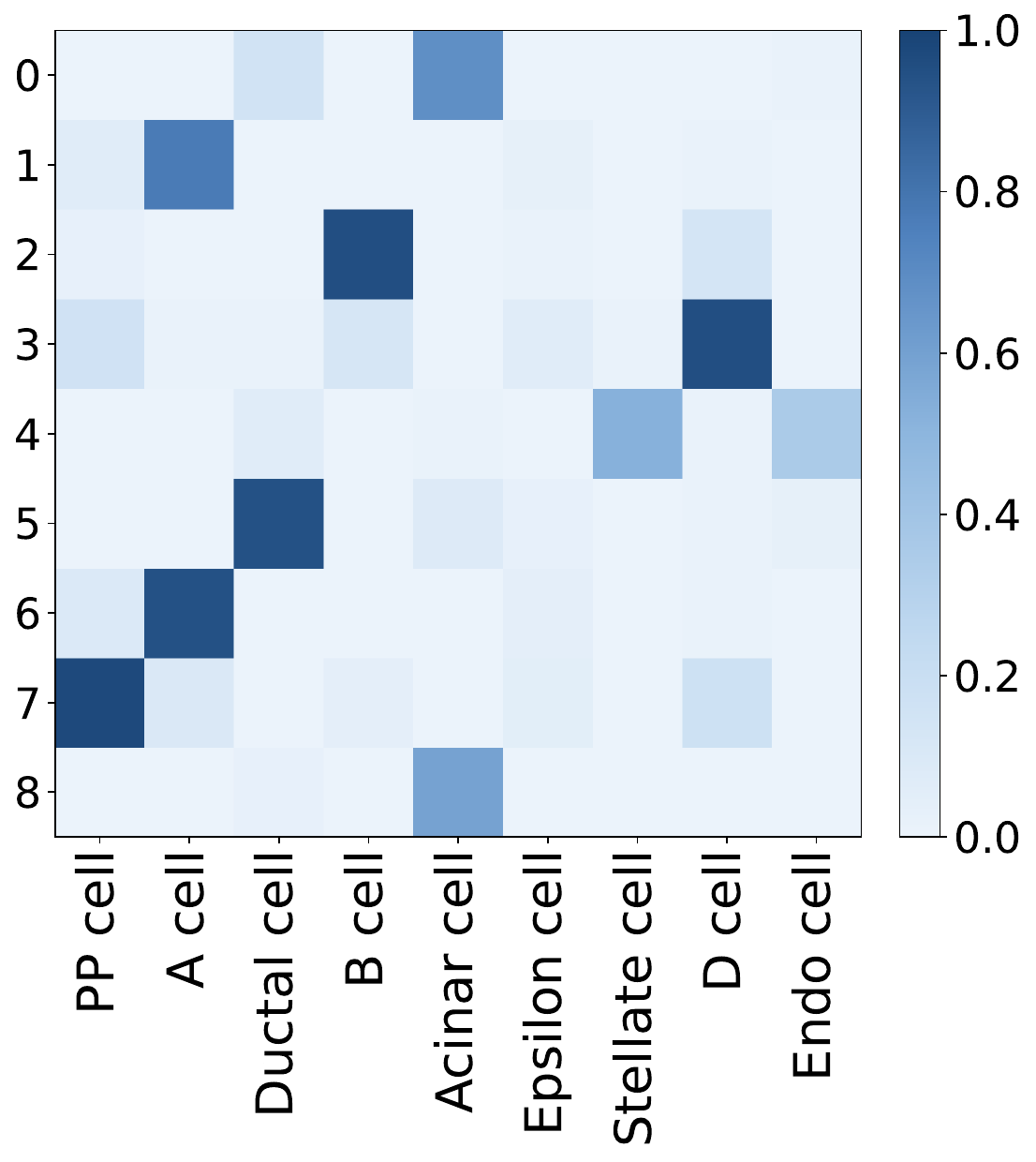}
        \caption{scCDCG}
    \end{subfigure}\hfill
    \begin{subfigure}[b]{0.19\linewidth}
        \includegraphics[width=\linewidth]{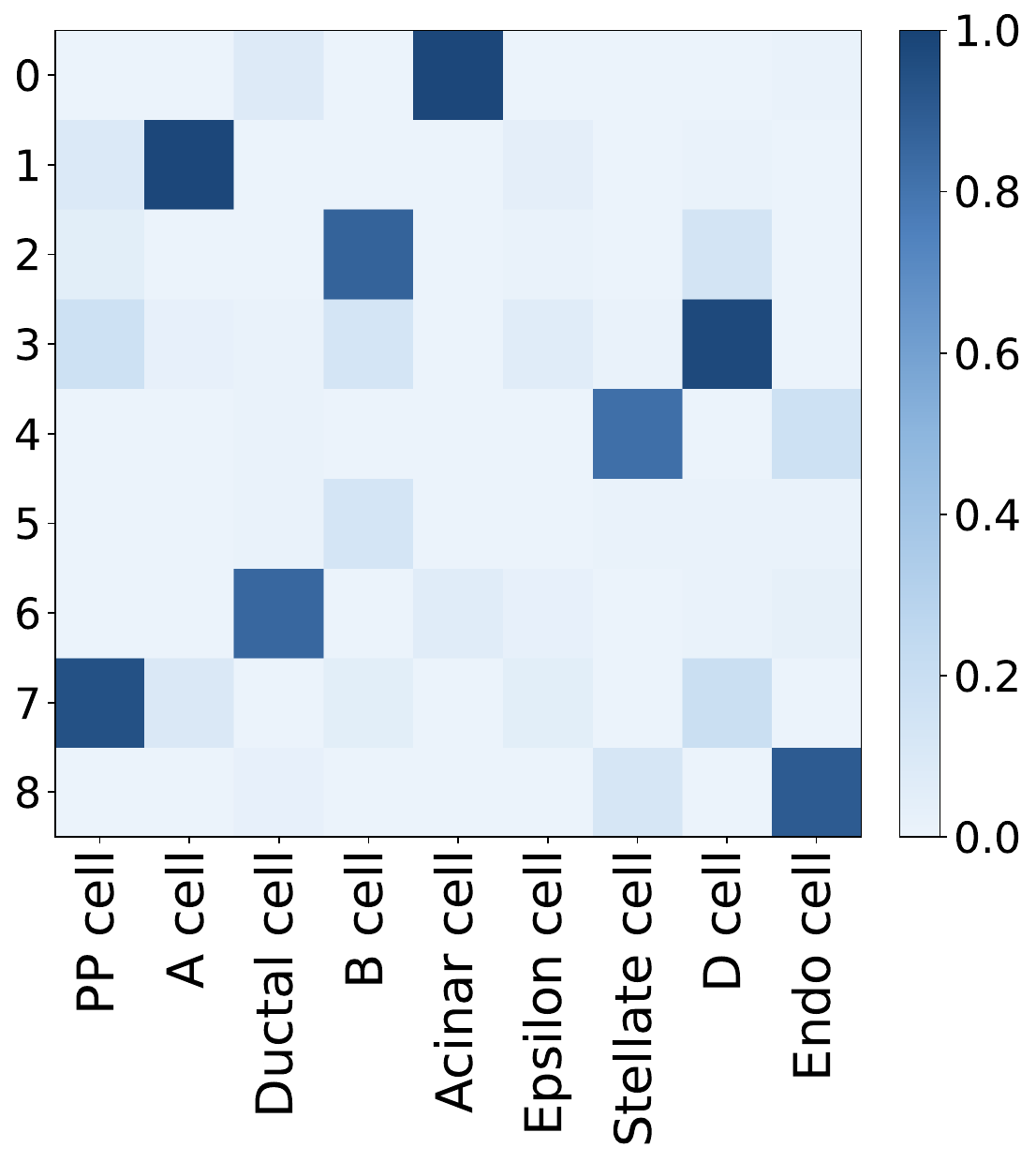}
        \caption{scSiameseClu}
    \end{subfigure}\hfill
    \begin{subfigure}[b]{0.19\linewidth}
        \includegraphics[width=\linewidth]{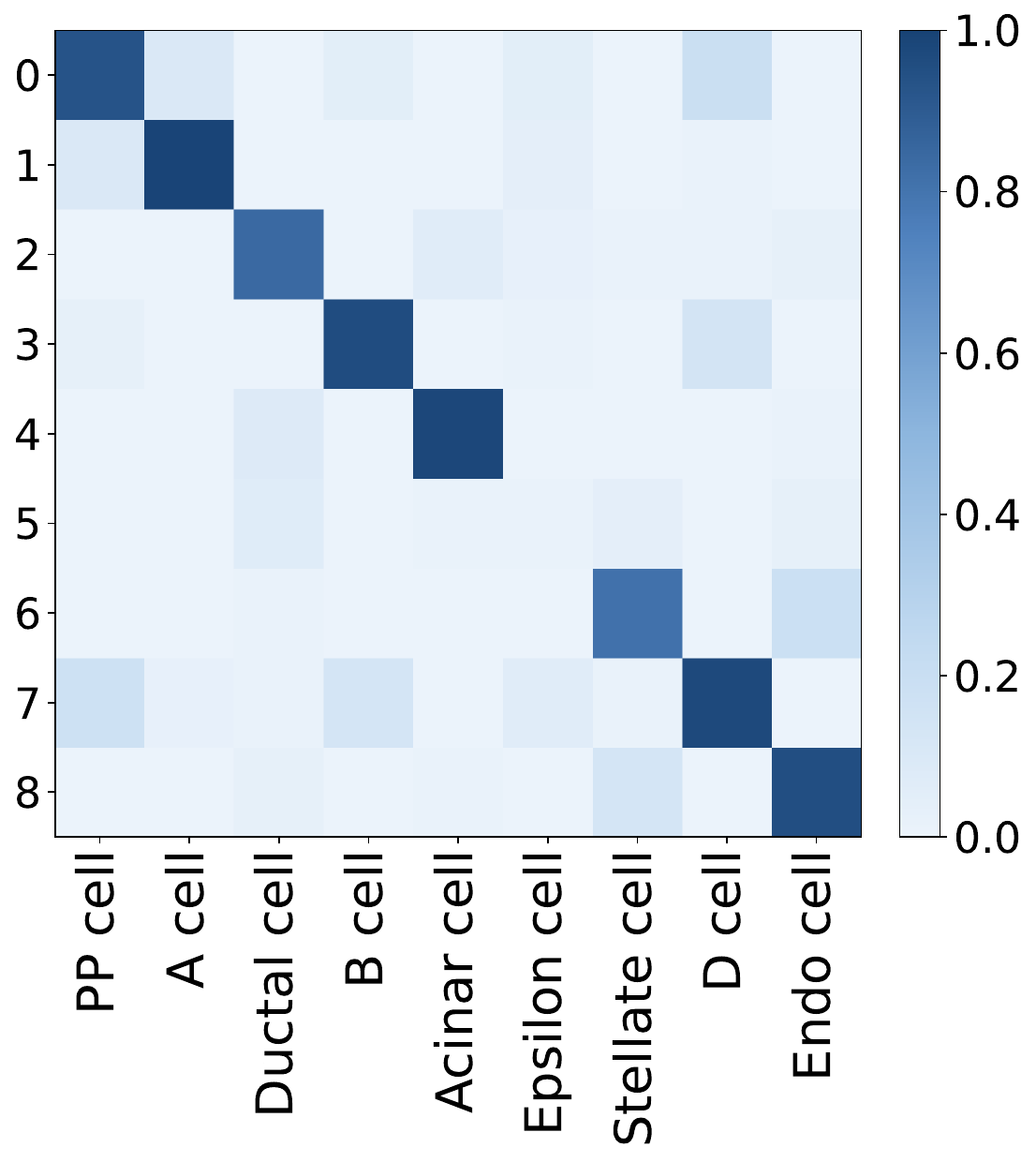}
        \caption{\method{}}
    \end{subfigure}
    \caption{Cell type annotation. Heatmap of overlap between the top 100 DEGs in clusters detected by five methods and gold standard cell types (Cell types are abbreviated as PP, A, Ductal, B, Acinar, Epsilon, Stellate, D, and Endo).}
    \label{fig:cell_type_annotation_heatmap}
\end{figure*}

\begin{figure*}[t]
  \centering
  \begin{subfigure}[t]{0.48\textwidth}
    \centering
    \includegraphics[width=\linewidth]{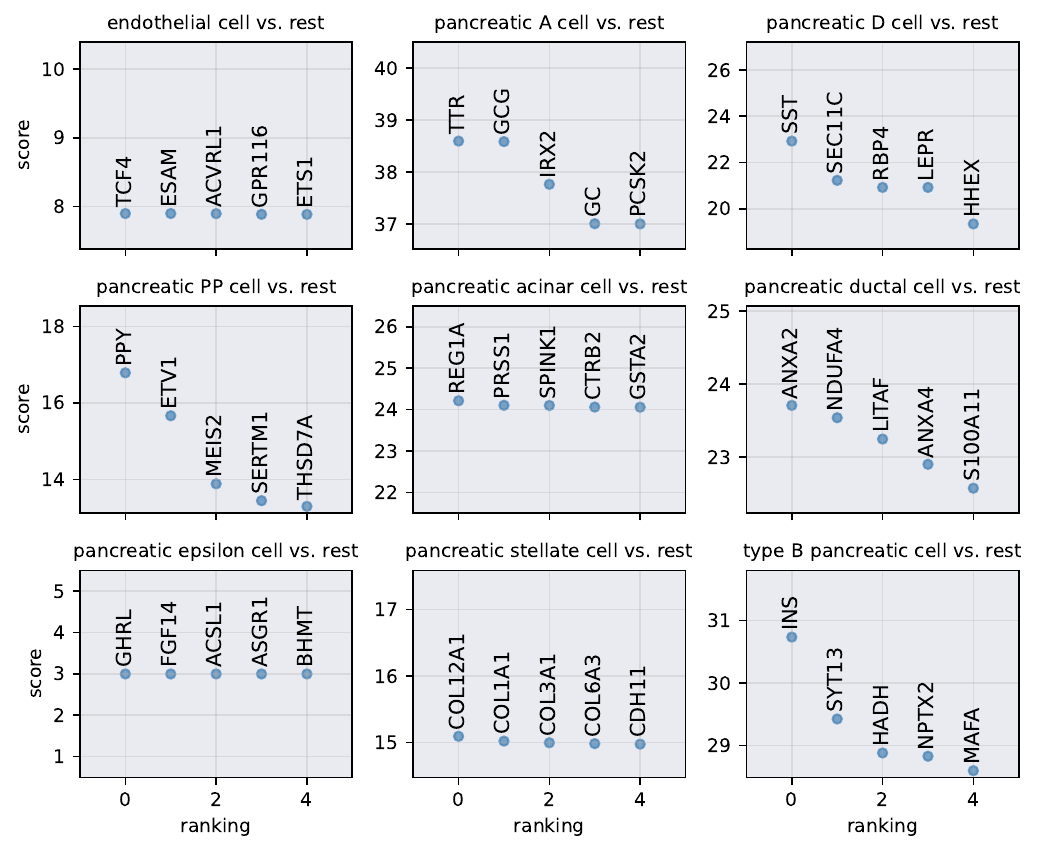}
    \caption{Gold Standard}
    \label{fig:deg_gold_10}
  \end{subfigure}
  \hfill
  \begin{subfigure}[t]{0.48\textwidth}
    \centering
    \includegraphics[width=\linewidth]{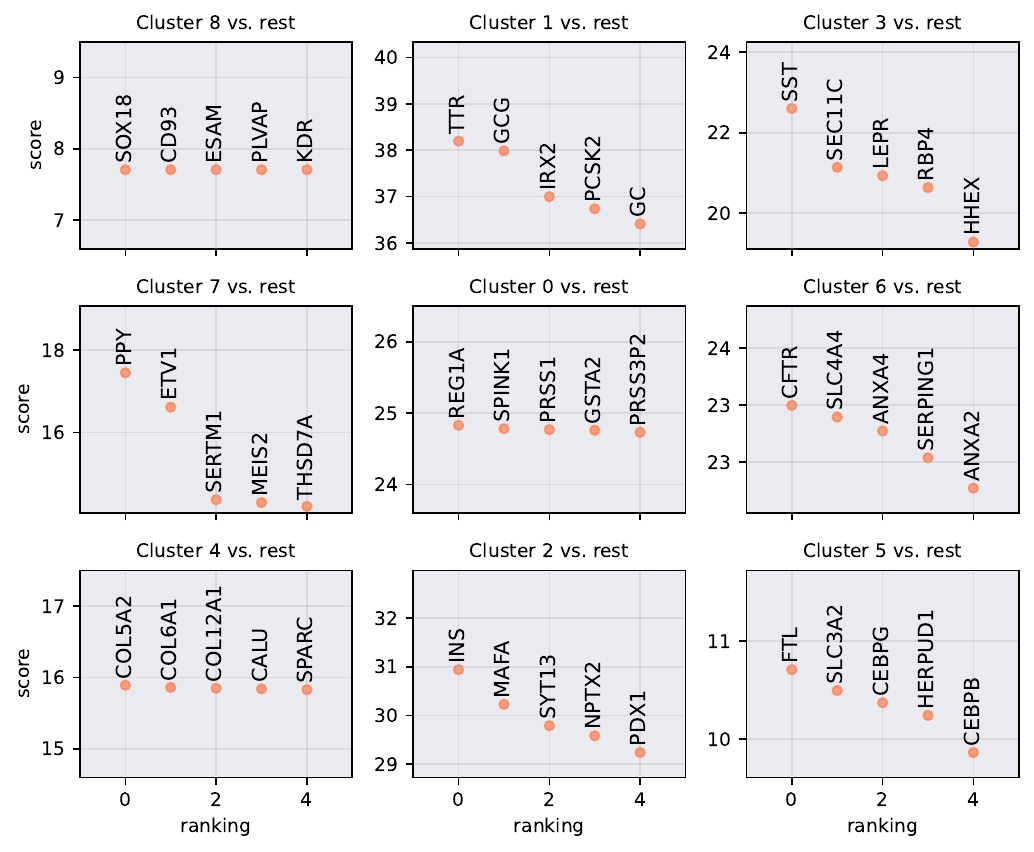}
    \caption{\method{}}
    \label{fig:deg_ours_10}
  \end{subfigure}

  \caption{Top-5 DEGs Ranking. Comparison of the top 5 DEGs identified by the gold standard (a) and \method{} (b) for different cell types.}
  \label{fig:deg_ranking_comparison}
\end{figure*}

\subsection{Overall Performance}

\smallskip\noindent\textbf{Quantitative Analysis}. Table~\ref{tab:main_result} compares the clustering performance of \method{} with ten competitive baselines on seven benchmark datasets. From the results, we make the following observations. 
(1) \method{} achieves the best performance across all three metrics on all datasets, demonstrating its strong capability in handling high-dimensional and highly sparse scRNA-seq data. On average, compared with the second-best method, \method{} yields significant improvements of \textbf{3.00\%}, \textbf{2.42\%}, and \textbf{2.30\%} in terms of ACC, NMI, and ARI, respectively, validating the effectiveness and generalizability of our approach.
(2) Compared with deep clustering methods that rely solely on gene expression features, our \method{} consistently delivers substantial gains, highlighting the benefit of jointly leveraging expression features and cell--cell structural information.
(3) Although GNN-based methods like scGNN incorporate cell graphs, they predominantly rely on local neighborhood aggregation, which renders them vulnerable to over-smoothing and limits their ability to capture global dependencies. \method{} overcomes this by encoding complex structural information via the Siamese Graph Transformer, yielding more discriminative representations. Analysis of runtime is provided in Appendix~\ref{sec:runtime}.

\smallskip\noindent\textbf{Visualization}. To further investigate the model's ability to capture the underlying structure of cells in the embedding space, we apply UMAP~\cite{mcinnes2020umap} to project the learned representations of the Human Liver dataset into two dimensions, as shown in Figure~\ref{fig:UMAP_comparison}. We observe that our \method{} produces highly compact clusters with clear inter-cluster boundaries, faithfully reflecting the complex underlying tissue structure and effectively preserving the intrinsic heterogeneity among different cell types. In contrast, methods such as scDSC often yield projections with noticeable subtype mixing or diffuse distributions, making it difficult to separate rare cell populations from major groups.

\subsection{Biological Analysis}
\smallskip\noindent\textbf{Differential Gene Expression Analysis}. 
Differentially expressed genes (DEGs) serve as a critical metric to validate the biological identity of learned clusters. Using Seurat's ``FindAllMarkers''~\cite{butler2018integrating}, we extract the top-100 DEGs for each cluster on the Muraro Human Pancreas dataset and quantitatively compute the overlap ratio with the corresponding ground-truth cell types. As shown in Figure~\ref{fig:cell_type_annotation_heatmap}, the high-overlap regions of \method{} are predominantly concentrated along the diagonal, consistently surpassing 90\% overlap in most clusters. This indicates that our clusters can be stably mapped to specific cell types with minimal confusion. In stark contrast, baseline methods frequently exhibit lower overlap ratios due to ambiguity and inter-cluster confusion. To further investigate the biological fidelity, we compare the rankings of the top 5 DEGs between \method{} and the Gold Standard as illustrated in Figure~\ref{fig:deg_ranking_comparison}. Beyond mere overlap, we observe remarkable consistency in the dominant markers. For instance, in the cluster corresponding to Pancreatic PP cells (Cluster 7), our \method{} identifies \textit{PPY} and \textit{ETV1} as the top-two markers, perfectly matching the Gold Standard. This precise alignment in both gene identity and ranking confidence demonstrates that our \method{} does not merely capture global similarities but accurately prioritizes the key marker genes essential for distinct cell identification.

\begin{figure*}[t]
    \centering
    \begin{subfigure}[b]{0.32\linewidth}
        \includegraphics[width=\linewidth]{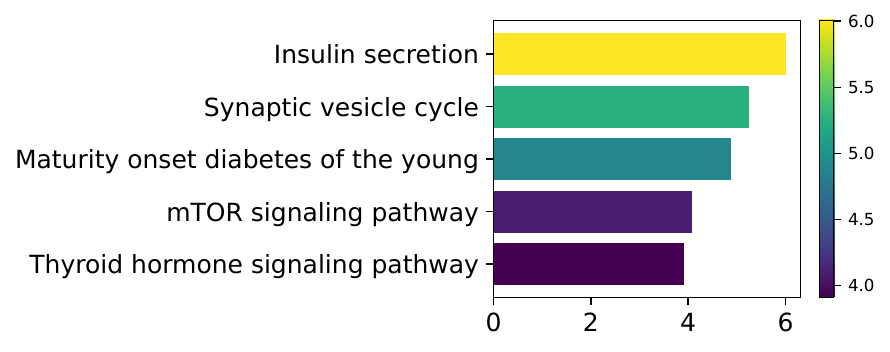}
        \caption{Cluster 3}
    \end{subfigure}\hfill
    \begin{subfigure}[b]{0.32\linewidth}
        \includegraphics[width=\linewidth]{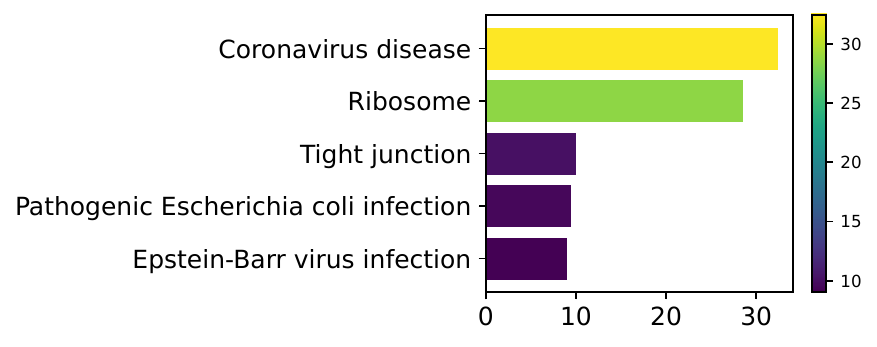}
        \caption{Cluster 4}
    \end{subfigure}\hfill
    \begin{subfigure}[b]{0.32\linewidth}
        \includegraphics[width=\linewidth]{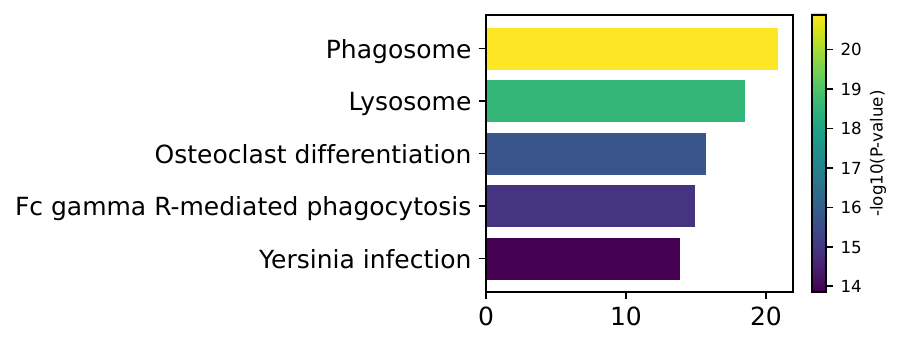}
        \caption{Cluster 5}
    \end{subfigure}
    \caption{Top 5 KEGG pathways for each cluster (bar graph shows $-\log_{10}(q)$).}
    \label{fig:kegg_bar}
\end{figure*}

\smallskip\noindent\textbf{KEGG Pathway Enrichment Analysis}. To validate the biological interpretability of the clustering results, we perform KEGG pathway enrichment analysis on representative clusters. As shown in Figure~\ref{fig:kegg_bar}, our \method{} successfully resolves distinct functional subpopulations in pancreatic tissue. Cluster 3 is significantly enriched in ``Insulin secretion'' and critical diabetes-related pathways including MODY, while also exhibiting neuroendocrine signatures such as the ``Synaptic vesicle cycle'', confirming its biological identity as pancreatic $\beta$ cells. Cluster 4 shows strong enrichment in ``Ribosome'' and protein synthesis-related pathways, highlighting the high protein synthesis activity characteristic of acinar cells. Finally, Cluster 5 is enriched in pathways such as ``Phagosome'' and ``Lysosome'', indicating the presence of tissue-resident immune cells such as macrophages. These results demonstrate that our \method{} can accurately identify KEGG pathways and distinct functional phenotypes.

Furthermore, we provide trajectory inference analysis in Appendix~\ref{sec:trajectory} to validate preserved developmental manifolds.

 \begin{figure}[t]
    \centering
    \begin{subfigure}[b]{0.48\linewidth}
        \includegraphics[width=\linewidth]{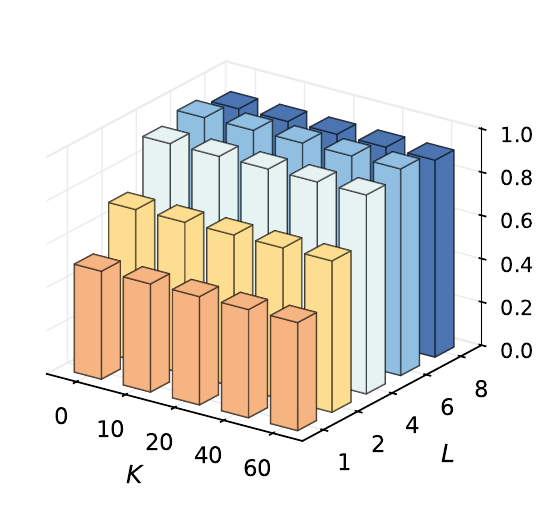}
        \caption{ACC}
    \end{subfigure}
    \hfill
    \begin{subfigure}[b]{0.48\linewidth}
        \includegraphics[width=\linewidth]{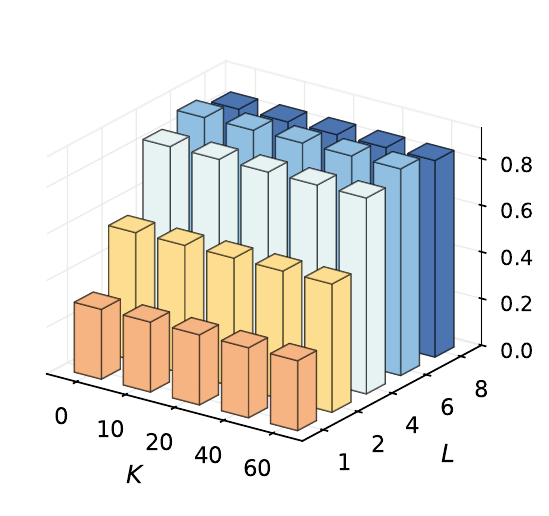}
        \caption{NMI}
    \end{subfigure}
    \caption{The impact of two key hyperparameters ($K$ and $L$) on clustering performance in terms of ACC and NMI.}
    \label{fig:sensitivity}
\end{figure}
\subsection{Ablation Study}

\smallskip\noindent\textbf{Effectiveness of Major Components.}
We evaluate the contribution of each loss term on the Muraro dataset. Table~\ref{tab:main_ablation} indicates that the exclusion of any component leads to a sharp performance degradation of over 10\% in ACC, highlighting the non-redundancy of each module. Specifically, the absence of the optimal transport clustering loss $\mathcal{L}_{\text{clu}}$ results in the most severe drop, underscoring its critical role in guiding partition assignments. Similarly, removing the correlation term $\mathcal{L}_{\text{cor}}$ or the reconstruction objectives $\mathcal{L}_{\text{ZINB}}$ and $\mathcal{L}_{\text{rec}}$ significantly impairs the capability of the model to align views and preserve topology. Ultimately, the full \method{} achieves superior performance, validating the synergy of jointly optimizing for structural, feature, and distribution constraints. Dual-component ablation analysis is provided in Appendix~\ref{sec:extended_ablation}.
\begin{table}[t]
\centering

\small
\begin{tabular}{cccc|ccc}
\toprule
\boldmath$\mathcal{L}_{\text{clu}}$ & \boldmath$\mathcal{L}_{\text{cor}}$ & \boldmath$\mathcal{L}_{\text{ZINB}}$ & \boldmath$\mathcal{L}_{\text{rec}}$ & \textbf{ACC} & \textbf{NMI} & \textbf{ARI} \\
\midrule
$\checkmark$ & $\checkmark$ & $\checkmark$ & $\times$      & 84.57 & 78.93 & 76.08 \\
$\checkmark$ & $\checkmark$ & $\times$      & $\checkmark$ & 84.18 & 78.36 & 75.62 \\
$\checkmark$ & $\times$      & $\checkmark$ & $\checkmark$ & 83.47 & 77.58 & 74.36 \\
$\times$      & $\checkmark$ & $\checkmark$ & $\checkmark$ & 81.24 & 74.79 & 70.68 \\
\midrule
$\checkmark$ & $\checkmark$ & $\checkmark$ & $\checkmark$ & \textbf{96.02} & \textbf{89.15} & \textbf{93.10} \\
\bottomrule
\end{tabular}
\caption{Ablation experiment results of major loss terms}
\label{tab:main_ablation}
\end{table}

\smallskip\noindent\textbf{Impact of Embedding Components.} We further dissect the embedding module to assess the contributions of Gene Expression (GE), Position (Pos), and Shortest-Path (SP) embeddings in Table~\ref{tab:ablation_embed}. Results indicate that while GE provides the foundational semantics, integrating it with structural priors yields massive gains. Notably, adding Pos alone boosts ACC by over 12\% compared to the GE baseline, and the full combination achieves peak performance, confirming that structural information is essential for distinguishing subtle cell states.
\begin{table}[t]
\centering

\small
\begin{tabular}{lccc}
\toprule
\textbf{Model Variants} & \textbf{ACC} & \textbf{NMI} & \textbf{ARI} \\
\midrule
\textbf{\method{}} & \textbf{96.02} & \textbf{89.15} & \textbf{93.10} \\
GE + Pos & 92.12 & 86.33 & 88.55 \\
GE + SP & 80.09 & 71.08 & 67.76 \\
Pos + SP & 27.52 & 12.13 & 5.06 \\
\midrule
Gene Expression (GE) & 79.24 & 70.69 & 67.61 \\
Position (Pos) & 16.68 & 0.73 & 0.23 \\
Shortest Path (SP) & 26.53 & 12.14 & 6.80 \\
\bottomrule
\end{tabular}
\caption{Ablation experiment results of embedding components.}
\label{tab:ablation_embed}
\end{table}

\subsection{Sensitivity Analysis}
To evaluate the robustness of \method{} against hyperparameter variations, we conduct a sensitivity analysis on the Muraro Human Pancreas dataset. We examine neighborhood size $K$ (5--60) and Transformer layers $L$ (1--8). As illustrated in Figure~\ref{fig:sensitivity}, the model demonstrates high stability with respect to $K$, showing only marginal fluctuations across all metrics, indicating that our \method{} is robust to variations in the underlying graph construction. Conversely, performance depends more significantly on network depth $L$. Shallow networks where $L=1$ yield suboptimal scores due to insufficient feature propagation and limited semantic capture. Increasing depth improves results until peaking at $L=6$, where the model effectively balances local and global information. Further extending depth to 8 leads to a slight performance drop, likely due to over-smoothing. Further sensitivity analyses regarding other hyperparameters are detailed in Appendix~\ref{sec:hyperparameter}.
\section{Conclusion}
In this paper, we propose a novel deep Siamese Graph Transformer Network (\method{}), which effectively integrates gene expression profiles and intercellular structural information for scRNA-seq data clustering. Specifically, we formulate the scRNA-seq data as a graph and construct two augmented graphs as dual views to capture complementary intercellular relationships. Then, we introduce a Siamese graph transformer network which integrates richer graph structural information, i.e., shortest-path and node-wise distance, to explicitly capture intercellular relationships. To further refine the cell clustering, we incorporate an optimal transport mechanism in a self-supervised manner, enhancing representation learning while preserving biological relevance. Extensive experiments on scRNA-seq benchmark datasets verify the effectiveness and superiority of our proposed \method{}.



\section*{Contribution Statement}
Jinke Wu and Yifan Wang contributed equally to this work.

\section*{Acknowledgments}
This work is supported in part by the National Natural Science Foundation of China under Grant 62306014 and 12501344, Postdoctoral Fellowship Program (Grade A) of CPSF under Grant BX20250376 and BX20240239, China Postdoctoral Science Foundation under Grant 2024M762201, Sichuan Science and Technology Program under Grant 2025ZNSFSC1506 and 2025ZNSFSC0808, Sichuan University Interdisciplinary Innovation Fund.

\bibliographystyle{named}
\bibliography{ijcai26}

@String { ARXIV        = {arXiv} }

@String { AAAI         = {Proceedings of the AAAI Conference on Artificial Intelligence}}

@String { IJCAI         = {Proceedings of the International Joint Conference on Artificial Intelligence}}

@String { WWW        = {Proceedings of the Web Conference}}

@article{peng2020single,
  title={Single-cell RNA-seq clustering: datasets, models, and algorithms},
  author={Peng, Lihong and Tian, Xiongfei and Tian, Geng and Xu, Junlin and Huang, Xin and Weng, Yanbin and Yang, Jialiang and Zhou, Liqian},
  journal={RNA biology},
  volume={17},
  number={6},
  pages={765--783},
  year={2020},
  publisher={Taylor \& Francis}
}

@article{kiselev2019challenges,
  title={Challenges in unsupervised clustering of single-cell RNA-seq data},
  author={Kiselev, Vladimir Yu and Andrews, Tallulah S and Hemberg, Martin},
  journal={Nature Reviews Genetics},
  volume={20},
  number={5},
  pages={273--282},
  year={2019},
  publisher={Nature Publishing Group UK London}
}

@inproceedings{xu2025scsiameseclu,
  title={scsiameseclu: A siamese clustering framework for interpreting single-cell rna sequencing data},
  author={Xu, Ping and Ning, Zhiyuan and Li, Pengjiang and Liu, Wenhao and Wang, Pengyang and Cui, Jiaxu and Zhou, Yuanchun and Wang, Pengfei},
  booktitle=IJCAI,
  pages={7867--7875},
  year={2025}
}

@article{hu2020iterative,
  title={Iterative transfer learning with neural network for clustering and cell type classification in single-cell RNA-seq analysis},
  author={Hu, Jian and Li, Xiangjie and Hu, Gang and Lyu, Yafei and Susztak, Katalin and Li, Mingyao},
  journal={Nature machine intelligence},
  volume={2},
  number={10},
  pages={607--618},
  year={2020},
  publisher={Nature Publishing Group UK London}
}

@article{dai2022accurate,
  title={Accurate and fast cell marker gene identification with COSG},
  author={Dai, Min and Pei, Xiaobing and Wang, Xiu-Jie},
  journal={Briefings in bioinformatics},
  volume={23},
  number={2},
  pages={bbab579},
  year={2022},
  publisher={Oxford University Press}
}

@article{hicks2021mbkmeans,
  title={mbkmeans: Fast clustering for single cell data using mini-batch k-means},
  author={Hicks, Stephanie C and Liu, Ruoxi and Ni, Yuwei and Purdom, Elizabeth and Risso, Davide},
  journal={PLoS computational biology},
  volume={17},
  number={1},
  pages={e1008625},
  year={2021},
  publisher={Public Library of Science San Francisco, CA USA}
}

@article{petegrosso2020machine,
  title={Machine learning and statistical methods for clustering single-cell RNA-sequencing data},
  author={Petegrosso, Raphael and Li, Zhuliu and Kuang, Rui},
  journal={Briefings in bioinformatics},
  volume={21},
  number={4},
  pages={1209--1223},
  year={2020},
  publisher={Oxford University Press}
}

@article{wang2021scgnn,
  title={scGNN is a novel graph neural network framework for single-cell RNA-Seq analyses},
  author={Wang, Juexin and Ma, Anjun and Chang, Yuzhou and Gong, Jianting and Jiang, Yuexu and Qi, Ren and Wang, Cankun and Fu, Hongjun and Ma, Qin and Xu, Dong},
  journal={Nature communications},
  volume={12},
  number={1},
  pages={1882},
  year={2021},
  publisher={Nature Publishing Group UK London}
}

@article{gan2022deep,
  title={Deep structural clustering for single-cell RNA-seq data jointly through autoencoder and graph neural network},
  author={Gan, Yanglan and Huang, Xingyu and Zou, Guobing and Zhou, Shuigeng and Guan, Jihong},
  journal={Briefings in Bioinformatics},
  volume={23},
  number={2},
  year={2022},
  publisher={Oxford Academic}
}

@article{johnson1967hierarchical,
  title={Hierarchical clustering schemes},
  author={Johnson, Stephen C},
  journal={Psychometrika},
  volume={32},
  number={3},
  pages={241--254},
  year={1967},
  publisher={Springer-Verlag}
}

@inproceedings{xu2024sccdcg,
  title={scCDCG: efficient deep structural clustering for single-cell RNA-Seq via deep cut-informed graph embedding},
  author={Xu, Ping and Ning, Zhiyuan and Xiao, Meng and Feng, Guihai and Li, Xin and Zhou, Yuanchun and Wang, Pengfei},
  booktitle={International Conference on Database Systems for Advanced Applications},
  pages={172--187},
  year={2024},
  organization={Springer}
}

@article{ou2022matrix,
  title={Matrix factorization for biomedical link prediction and scRNA-seq data imputation: an empirical survey},
  author={Ou-Yang, Le and Lu, Fan and Zhang, Zi-Chao and Wu, Min},
  journal={Briefings in Bioinformatics},
  volume={23},
  number={1},
  pages={bbab479},
  year={2022},
  publisher={Oxford University Press}
}

@article{wu2022network,
  title={Network-based structural learning nonnegative matrix factorization algorithm for clustering of scRNA-seq data},
  author={Wu, Wenming and Ma, Xiaoke},
  journal={IEEE/ACM transactions on computational biology and bioinformatics},
  volume={20},
  number={1},
  pages={566--575},
  year={2022},
  publisher={IEEE}
}

@incollection{benesty2009pearson,
  title={Pearson correlation coefficient},
  author={Benesty, Jacob and Chen, Jingdong and Huang, Yiteng and Cohen, Israel},
  booktitle={Noise reduction in speech processing},
  pages={1--4},
  year={2009},
  publisher={Springer}
}

@article{maaten2008visualizing,
  title={Visualizing data using t-SNE},
  author={Maaten, Laurens van der and Hinton, Geoffrey},
  journal={Journal of machine learning research},
  volume={9},
  number={Nov},
  pages={2579--2605},
  year={2008}
}

@inproceedings{bo2020structural,
  title={Structural deep clustering network},
  author={Bo, Deyu and Wang, Xiao and Shi, Chuan and Zhu, Meiqi and Lu, Emiao and Cui, Peng},
  booktitle=WWW,
  pages={1400--1410},
  year={2020}
}

@article{sinkhorn1967diagonal,
  title={Diagonal equivalence to matrices with prescribed row and column sums},
  author={Sinkhorn, Richard},
  journal={The American Mathematical Monthly},
  volume={74},
  number={4},
  pages={402--405},
  year={1967},
  publisher={JSTOR}
}

@article{eraslan2019single,
  title={Single-cell RNA-seq denoising using a deep count autoencoder},
  author={Eraslan, G{\"o}kcen and Simon, Lukas M and Mircea, Maria and Mueller, Nikola S and Theis, Fabian J},
  journal={Nature communications},
  volume={10},
  number={1},
  pages={390},
  year={2019},
  publisher={Nature Publishing Group UK London}
}

@article{lin2017cidr,
  title={CIDR: Ultrafast and accurate clustering through imputation for single-cell RNA-seq data},
  author={Lin, Peijie and Troup, Michael and Ho, Joshua WK},
  journal={Genome biology},
  volume={18},
  number={1},
  pages={59},
  year={2017},
  publisher={Springer}
}

@article{stegle2015computational,
  title={Computational and analytical challenges in single-cell transcriptomics},
  author={Stegle, Oliver and Teichmann, Sarah A and Marioni, John C},
  journal={Nature Reviews Genetics},
  volume={16},
  number={3},
  pages={133--145},
  year={2015},
  publisher={Nature Publishing Group UK London}
}

@article{ciortan2021contrastive,
  title={Contrastive self-supervised clustering of scRNA-seq data},
  author={Ciortan, Madalina and Defrance, Matthieu},
  journal={BMC bioinformatics},
  volume={22},
  number={1},
  pages={280},
  year={2021},
  publisher={Springer}
}

@inproceedings{wang2025deep,
  title={Deep Multi-modal Graph Clustering via Graph Transformer Network},
  author={Wang, Qianqian and Xu, Haiming and Zhang, Zihao and Feng, Wei and Gao, Quanxue},
  booktitle={Proceedings of the AAAI Conference on Artificial Intelligence},
  volume={39},
  number={8},
  pages={7835--7843},
  year={2025}
}

@inproceedings{yu2022zinb,
  title={Zinb-based graph embedding autoencoder for single-cell rna-seq interpretations},
  author={Yu, Zhuohan and Lu, Yifu and Wang, Yunhe and Tang, Fan and Wong, Ka-Chun and Li, Xiangtao},
  booktitle={Proceedings of the AAAI conference on artificial intelligence},
  volume={36},
  number={4},
  pages={4671--4679},
  year={2022}
}

@inproceedings{li2018deeper,
  title={Deeper insights into graph convolutional networks for semi-supervised learning},
  author={Li, Qimai and Han, Zhichao and Wu, Xiao-Ming},
  booktitle={Proceedings of the AAAI conference on artificial intelligence},
  volume={32},
  number={1},
  year={2018}
}

@article{vzurauskiene2016pcareduce,
  title={pcaReduce: hierarchical clustering of single cell transcriptional profiles},
  author={{\v{Z}}urauskien{\.e}, Justina and Yau, Christopher},
  journal={BMC bioinformatics},
  volume={17},
  number={1},
  pages={140},
  year={2016},
  publisher={Springer}
}

@article{strehl-ghosh:cluster-ensembles,
  author  = "Alexander Strehl and Joydeep Ghosh",
  title   = "Cluster Ensembles---A Knowledge Reuse Framework for Combining Multiple Partitions",
  journal = "Journal of Machine Learning Research",
  volume  = "3",
  pages   = "583--617",
  year    = "2002"
}

@inproceedings{vinh-et-al:ari,
  author    = "Nguyen Xuan Vinh and Julien Epps and James Bailey",
  title     = "Information Theoretic Measures for Clusterings Comparison: Is a Correction for Chance Necessary?",
  booktitle = "Proceedings of the 26th Annual International Conference on Machine Learning",
  pages     = "1073--1080",
  year      = "2009"
}

@inproceedings{xie-et-al:dec,
  author    = "Junyuan Xie and Ross Girshick and Ali Farhadi",
  title     = "Unsupervised Deep Embedding for Clustering Analysis",
  booktitle = "Proceedings of the 33rd International Conference on Machine Learning",
  pages     = "478--487",
  year      = "2016"
}

@article{tian-et-al:model-based-scrnaseq,
  author  = "Tian Tian and Ji Wan and Qi Song and Zhi Wei",
  title   = "Clustering Single-Cell {RNA}-seq Data with a Model-Based Deep Learning Approach",
  journal = "Nature Machine Intelligence",
  volume  = "1",
  number  = "4",
  pages   = "191--198",
  year    = "2019"
}

@article{wan-et-al:scname,
  author  = "Hui Wan and Liang Chen and Minghua Deng",
  title   = "{scNAME}: Neighborhood Contrastive Clustering with Ancillary Mask Estimation for {scRNA}-seq Data",
  journal = "Bioinformatics",
  volume  = "38",
  number  = "6",
  pages   = "1575--1583",
  year    = "2022"
}

@article{li-et-al:attention-scrnaseq,
  author  = "Shenghao Li and Hui Guo and Simai Zhang and Yizhou Li and Menglong Li",
  title   = "Attention-Based Deep Clustering Method for {scRNA}-seq Cell Type Identification",
  journal = "PLOS Computational Biology",
  volume  = "19",
  number  = "11",
  pages   = "e1011641",
  year    = "2023"
}

@article{tian2021model,
  title  = {Model-based deep embedding for constrained clustering analysis of {single cell RNA-seq} data},
  author = {Tian, Tian and Zhang, Jie and Lin, Xiang and Wei, Zhi and Hakonarson, Hakon},
  journal = {Nature Communications},
  volume = {12},
  number = {1},
  pages  = {1873},
  year   = {2021},
  publisher = {Nature Publishing Group}
}

@article{butler2018integrating,
  title  = {Integrating single-cell transcriptomic data across different conditions, technologies, and species},
  author = {Butler, Andrew and Hoffman, Paul and Smibert, Peter and Papalexi, Efthymia and Satija, Rahul},
  journal = {Nature Biotechnology},
  volume = {36},
  number = {5},
  pages  = {411--420},
  year   = {2018},
  publisher = {Nature Publishing Group}
}

@article{mcinnes2020umap,
  title  = {{UMAP}: Uniform Manifold Approximation and Projection for Dimension Reduction},
  author = {McInnes, Leland and Healy, John and Melville, James},
  journal = {arXiv preprint arXiv:1802.03426},
  year   = {2020}
}

@article{luo2021topology,
  title={A topology-preserving dimensionality reduction method for single-cell RNA-seq data using graph autoencoder},
  author={Luo, Zixiang and Xu, Chenyu and Zhang, Zhen and Jin, Wenfei},
  journal={Scientific reports},
  volume={11},
  number={1},
  pages={20028},
  year={2021},
  publisher={Nature Publishing Group UK London}
}

@article{wolf2019paga,
  title={PAGA: graph abstraction reconciles clustering with trajectory inference through a topology preserving map of single cells},
  author={Wolf, F Alexander and Hamey, Fiona K and Plass, Mireya and Solana, Jordi and Dahlin, Joakim S and G{\"o}ttgens, Berthold and Rajewsky, Nikolaus and Simon, Lukas and Theis, Fabian J},
  journal={Genome biology},
  volume={20},
  number={1},
  pages={59},
  year={2019},
  publisher={Springer}
}

@article{street2018slingshot,
  title={Slingshot: cell lineage and pseudotime inference for single-cell transcriptomics},
  author={Street, Kelly and Risso, Davide and Fletcher, Russell B and Das, Diya and Ngai, John and Yosef, Nir and Purdom, Elizabeth and Dudoit, Sandrine},
  journal={BMC genomics},
  volume={19},
  number={1},
  pages={477},
  year={2018},
  publisher={Springer}
}

@article{kingma2014adam,
  title={Adam: A method for stochastic optimization},
  author={Kingma, Diederik P},
  journal={arXiv preprint arXiv:1412.6980},
  year={2014}
}

@inproceedings{zhang2026evidence,
  title={Evidence-aware Integration and Domain Identification of Spatial Transcriptomics Data},
  author={Zhang, Wei and Yi, Siyu and Chen, Lezhi and Wang, Yifan and Qiao, Ziyue and Zhou, Yongdao and Ju, Wei},
  booktitle={Proceedings of the AAAI Conference on Artificial Intelligence},
  volume={40},
  number={19},
  pages={16352--16360},
  year={2026}
}

@article{ju2025survey,
  title={A survey of graph neural networks in real world: Imbalance, noise, privacy and ood challenges},
  author={Ju, Wei and Yi, Siyu and Wang, Yifan and Xiao, Zhiping and Mao, Zhengyang and Li, Hourun and Gu, Yiyang and Qin, Yifang and Yin, Nan and Wang, Senzhang and others},
  journal={IEEE Transactions on Pattern Analysis and Machine Intelligence},
  year={2025},
  publisher={IEEE}
}

@inproceedings{ju2023glcc,
  title={Glcc: A general framework for graph-level clustering},
  author={Ju, Wei and Gu, Yiyang and Chen, Binqi and Sun, Gongbo and Qin, Yifang and Liu, Xingyuming and Luo, Xiao and Zhang, Ming},
  booktitle={Proceedings of the AAAI Conference on Artificial Intelligence},
  volume={37},
  number={4},
  pages={4391--4399},
  year={2023}
}

@inproceedings{ju2026compactness,
  title={Compactness and Consistency: A Conjoint Framework for Deep Graph Clustering},
  author={Ju, Wei and Yi, Siyu and Zheng, Kangjie and Wang, Yifan and Qiao, Ziyue and Shen, Li and Zhou, Yongdao and Cao, Xiaochun and Lv, Jiancheng},
  booktitle={The Fourteenth International Conference on Learning Representations},
  year={2026}
}

@article{fan2026cmgl,
  title={CMGL: Confidence-guided Multi-omics Graph Learning for Cancer Subtype Classification},
  author={Fan, Boyang and Yin, Hengchuang and Yi, Siyu and Wang, Yifan and Li, Zhicheng and Zhou, Leijiyu and Lv, Jiancheng and Ju, Wei},
  journal={arXiv preprint arXiv:2604.24201},
  year={2026}
}

@article{ren2025mhgc,
  title={MHGC: Multi-scale hard sample mining for contrastive deep graph clustering},
  author={Ren, Tao and Zhang, Haodong and Wang, Yifan and Ju, Wei and Liu, Chengwu and Meng, Fanchun and Yi, Siyu and Luo, Xiao},
  journal={Information Processing \& Management},
  volume={62},
  number={4},
  pages={104084},
  year={2025},
  publisher={Elsevier}
}

@inproceedings{zhang2026fairgc,
  title={FairGC: Fostering Individual and Group Fairness for Deep Graph Clustering},
  author={Zhang, Haodong and Wang, Xinyue and Ren, Tao and Wang, Yifan and Yi, Siyu and Meng, Fanchun and Ma, Zeyu and Long, Qingqing and Ju, Wei},
  booktitle={Proceedings of the AAAI Conference on Artificial Intelligence},
  volume={40},
  number={33},
  pages={28194--28202},
  year={2026}
}

\newpage
\twocolumn \flushbottom \sloppy
\begin{appendix}

\section{Related Work}
\label{sec::related}

\subsection{Classical Clustering Methods for scRNA-seq}
Numerous single-cell clustering methods have been proposed in recent years. Early approaches typically follow a two-stage paradigm: first obtaining low-dimensional features via dimensionality reduction techniques, and subsequently applying classical algorithms for clustering, such as  $k$-means~\cite{hicks2021mbkmeans}, hierarchical clustering~\cite{johnson1967hierarchical}, and density-based methods~\cite{petegrosso2020machine}. To learn low-dimensional representations, some methods, such as pcaReduce~\cite{vzurauskiene2016pcareduce}, integrate Principal Component Analysis (PCA) with $k$-means to iteratively merge clusters. In contrast, CIDR~\cite{lin2017cidr} reduces dimensionality by constructing a dissimilarity matrix based on imputed gene expression values. Nonetheless, these methods lack the flexibility to model complex nonlinear relationships among cells~\cite{stegle2015computational}. To overcome the limitations, various deep clustering methods have been developed to learn more expressive representations. DEC~\cite{xie-et-al:dec} pioneers this direction by learning feature representations and cluster assignments within a unified framework. Furthermore, scDeepCluster~\cite{tian-et-al:model-based-scrnaseq} and scDCC~\cite{tian2021model} incorporate Zero-Inflated Negative Binomial (ZINB) autoencoders, whereas contrastive-sc~\cite{ciortan2021contrastive} leverages contrastive learning for robustness. Despite their success, these methods primarily focus on intra-cellular gene expression patterns and often neglect topological relationships among cells.

\subsection{Deep Graph Clustering for scRNA-seq}
Recent research has pivoted toward deep graph clustering frameworks based on Graph Neural Networks (GNNs)~\cite{ju2023glcc,wang2025deep,ren2025mhgc,zhang2026fairgc,ju2026compactness}. By representing cells as nodes and their interactions as edges, these methods effectively integrate gene expression profiles with topological structures via message-passing mechanisms to learn context-aware cell embeddings. For instance, scGNN~\cite{wang2021scgnn} employs GNNs and multi-modal autoencoders to aggregate cell-cell relationships, while scGAE~\cite{luo2021topology} utilizes a multitask-oriented graph autoencoder to simultaneously preserve feature and topological information. To better handle the prevalence of dropout events and technical noise, scTAG~\cite{yu2022zinb} integrates a ZINB model into a topology-adaptive graph convolutional autoencoder to learn low-dimensional latent representations in an end-to-end manner. Similarly, scDSC~\cite{gan2022deep} combines a ZINB-based autoencoder with GNN modules through a mutually supervised strategy, allowing the structural and attribute information to mutually refine the clustering results. Most recently, scSiamese~\cite{xu2025scsiameseclu} leverages a Siamese clustering framework to enhance representation robustness via cross-view consistency. However, these GNN-based approaches primarily focus on local neighborhood aggregation and often overlook richer structural information, which results in over-smoothed embeddings that fail to distinguish biologically distinct populations~\cite{li2018deeper}.

\section{Detailed Experimental Setup}
\label{sec:setup}
Our proposed \method{} is implemented using PyTorch and optimized on a NVIDIA RTX 4090 GPU. Details of the datasets are summarized in Table~\ref{tab:dataset_summary}. In the graph construction phase, the neighborhood size $K$ is set to 20. For the dual augmentation module, we set the edge dropping rate to 0.1 to remove spurious connections, and the diffusion coefficient $\eta$ in the graph diffusion view is set to 0.2. The training consists of two stages: the framework is first pre-trained for 200 epochs to initialize the feature embeddings, followed by 200 epochs of joint training for the clustering task to ensure convergence. The entire model is optimized using the Adam optimizer~\cite{kingma2014adam}. For the clustering, the centroid of the soft assignment probability is initialized by $k$-means and updated based on the embeddings. Regarding the specific embedding construction for the Siamese Graph Transformer, the input $\bm{Y}_i^* \in \mathbb{R}^{(K+1)\times d}$ is derived via three specific mapping functions: $\bm{Y}_i^* = \bm{E}_i^* + \bm{P}_i^* + \bm{H}_i^*$. Specifically, the gene expression mapping $\mathcal{F}_{ge}(\cdot)$ is implemented as a learnable linear layer that projects the GNNs-aggregated features into a hidden dimension of $d=128$. The position mapping $\mathcal{F}_{pos}(\cdot)$ utilizes an embedding lookup table to encode rank-based indices, where the central node is assigned 0 and neighbors are assigned 1 to $K$ based on similarity sorting. Similarly, the shortest-path mapping $\mathcal{F}_{sp}(\cdot)$ employs a separate embedding layer to transform BFS-calculated hop counts into dense vectors. Finally, the fused embeddings are regularized with a dropout rate of 0.1 before entering the Transformer layers.

\begin{table}
\centering
\resizebox{0.95\linewidth}{!}{
\begin{tabular}{l|ccc}
\toprule
\textbf{Baselinse} & \textbf{ACC} & \textbf{NMI} & \textbf{ARI} \\
\midrule

pcaReduce & $1.56 \times 10^{-2}$ & $1.56 \times 10^{-2}$ & $1.56 \times 10^{-2}$ \\
DEC & $1.56 \times 10^{-2}$ & $1.56 \times 10^{-2}$ & $1.56 \times 10^{-2}$ \\
contrastive-sc & $< 10^{-3}$ & $< 10^{-3}$ & $< 10^{-3}$ \\
scNAME & $< 10^{-3}$ & $< 10^{-3}$ & $< 10^{-3}$ \\
scDeepCluster & $< 10^{-3}$ & $< 10^{-3}$ & $< 10^{-3}$ \\
AttentionAE-sc & $< 10^{-3}$ & $< 10^{-3}$ & $< 10^{-3}$ \\
scDSC & $< 10^{-3}$ & $< 10^{-3}$ & $< 10^{-3}$ \\
scGNN & $< 10^{-3}$ & $< 10^{-3}$ & $< 10^{-3}$ \\
scCDCG & $1.56 \times 10^{-2}$ & $1.56 \times 10^{-2}$ & $0.08$ \\ 
scSiameseClu & $0.04$ & $0.06$ & $0.03$ \\ 
\bottomrule
\end{tabular}
}
\caption{Statistical significance of \method{} compared to baselines. Results with $p < 0.05$ are considered statistically significant.}
\label{tab:significance}
\end{table}
\section{Statistical Significance Analysis} To strictly verify performance improvements, we conducted significance analysis between \method{} and ten baselines across seven benchmark datasets using the Wilcoxon signed-rank test. As reported in Table~\ref{tab:significance}, our \method{} demonstrates statistical significance ($p < 0.05$) in 93.8\% of pairwise comparisons. Specifically, it achieves overwhelming dominance over early methods (e.g., pcaReduce) with $p$-values typically below $10^{-3}$. Against strong SOTA competitors like scSiameseClu, \method{} maintains a competitive edge; although isolated metrics show marginal significance due to dataset variance, \method{} consistently yields the highest mean performance averaged across the seven datasets, confirming its superior robustness across diverse scenarios.

\begin{table*}[t] 
\centering
\small
\setlength{\tabcolsep}{6pt} 

\resizebox{0.8\textwidth}{!}{%
\begin{tabular}{lcccllcc}
\toprule
\textbf{Dataset} & \textbf{Cells} & \textbf{Genes} & \textbf{Clusters} & \textbf{Organ} & \textbf{Seq. Method} & \textbf{Sparsity} & \textbf{HVGs} \\
\midrule
\textbf{Muraro human Pancreas cell} & 2122 & 19046 & 9  & Pancreas & CEL-seq      & 73.02\% & 2000 \\
\textbf{Human Pancreas cell 1}      & 1937 & 20125 & 14 & Pancreas & CEL-seq      & 90.44\% & 2000 \\
\textbf{Human Pancreas cell 2}      & 1724 & 20125 & 14 & Pancreas & CEL-seq      & 90.59\% & 2000 \\
\textbf{Human Pancreas cell 3}      & 3605 & 20125 & 14 & Pancreas & CEL-seq      & 91.30\% & 2000 \\
\textbf{Mouse Pancreas cell 2}      & 1064 & 14878 & 13 & Pancreas & CEL-seq      & 87.81\% & 1500 \\
\textbf{CITE-CBMC}                  & 8617 & 2000  & 15 & Blood    & 10X Genomics & 93.26\% & 2000 \\
\textbf{Human Liver cells}          & 8444 & 5000  & 11 & Liver    & 10X Genomics & 90.77\% & 1500 \\
\bottomrule
\end{tabular}%
}
\caption{Summary of the scRNA-seq datasets.}
\label{tab:dataset_summary}
\vspace{2pt}
\end{table*}

\begin{figure*}[t]
    \centering
    \includegraphics[width=0.8\linewidth]{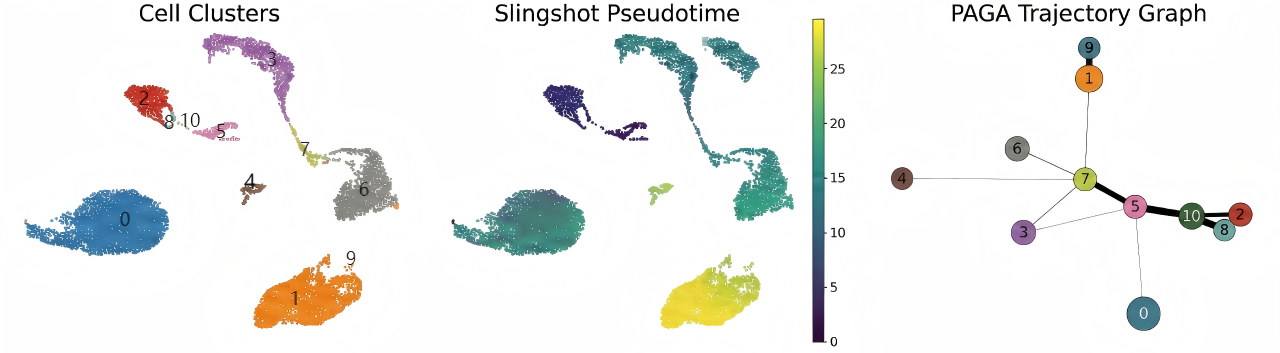} 
    \caption{Trajectory inference analysis on the Human Liver cells dataset.}
    \label{fig:trajectory}
\end{figure*}

\section{Trajectory Inference}
\label{sec:trajectory}
To substantiate the capability of \method{} in preserving global structural dependencies and the continuous developmental manifold, we perform trajectory inference on the Human Liver cells dataset using PAGA~\cite{wolf2019paga} and Slingshot~\cite{street2018slingshot}. As illustrated in Figure~\ref{fig:trajectory}, the PAGA graph constructed from \method{} embeddings effectively reconstructs a hierarchical lineage backbone originating from the progenitor populations located in Cluster 0 and Cluster 8. These clusters correspond to the regions with the earliest pseudotime states and exhibit strong connectivity to the transitional central hub identified as Cluster 2, which acts as a critical intermediate state bridging differentiation branches. Subsequently, the trajectory bifurcates into distinct functional lineages, such as the major differentiation path progressing through Clusters 1 and 9 towards the terminal states in Clusters 3 and 6. This topological structure is further corroborated by the continuous gradient observed in the Slingshot pseudotime visualization, confirming that \method{} successfully mitigates the over-smoothing issue typical of GNNs and captures the intrinsic non-linear manifold of cellular differentiation.


\section{Hyperparameter Sensitivity Analysis} 
\label{sec:hyperparameter}
\begin{figure}[t]
    \centering
    \includegraphics[width=0.6\linewidth]{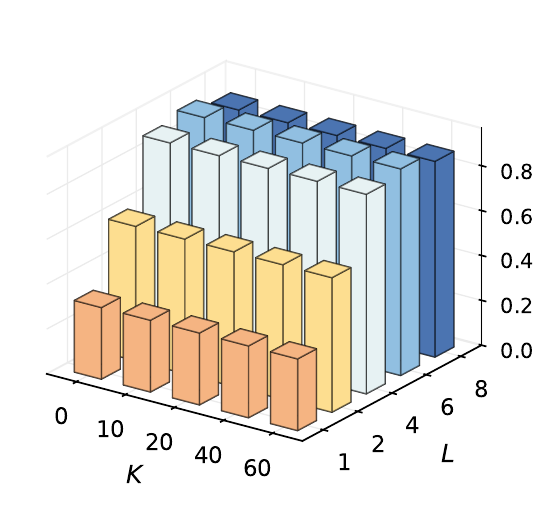}
    \caption{The impact of neighborhood size $K$ and the number of Transformer layers $L$ on clustering performance in terms of ARI on the Muraro Human Pancreas dataset.}
    \label{fig:sensitivity_ari}
\end{figure}

\begin{figure}
    \centering
    \begin{subfigure}[b]{0.49\linewidth}
        \centering
        \includegraphics[width=\linewidth]{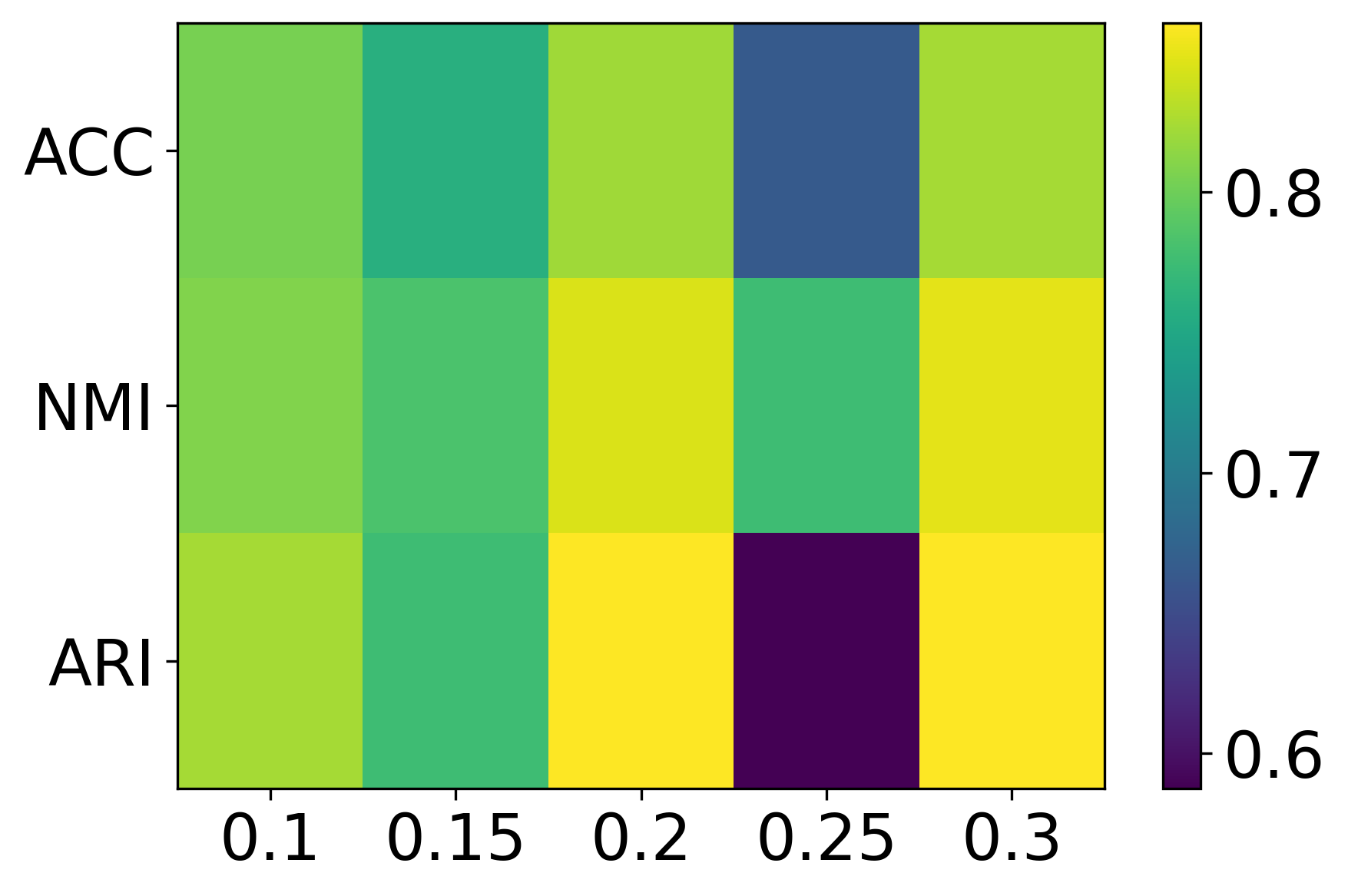} 
        \caption{$\eta$}
        \label{fig:sens_eta}
    \end{subfigure}
    \hfill 
    \begin{subfigure}[b]{0.49\linewidth}
        \centering
        \includegraphics[width=\linewidth]{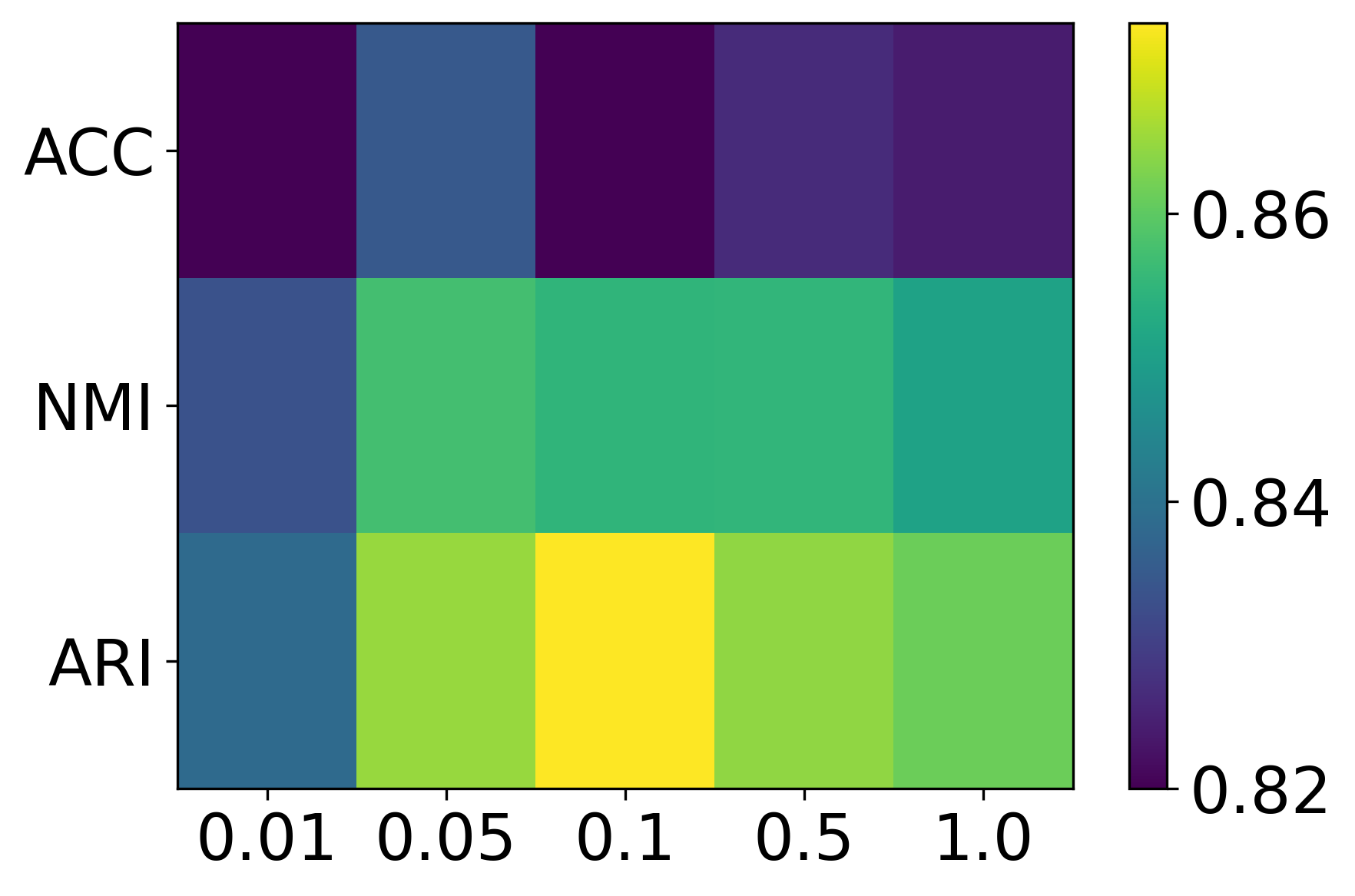}
        \caption{$\alpha$}
        \label{fig:sens_alpha}
    \end{subfigure}
    
    \vspace{0.2cm} 
    
    \begin{subfigure}[b]{0.49\linewidth}
        \centering
        \includegraphics[width=\linewidth]{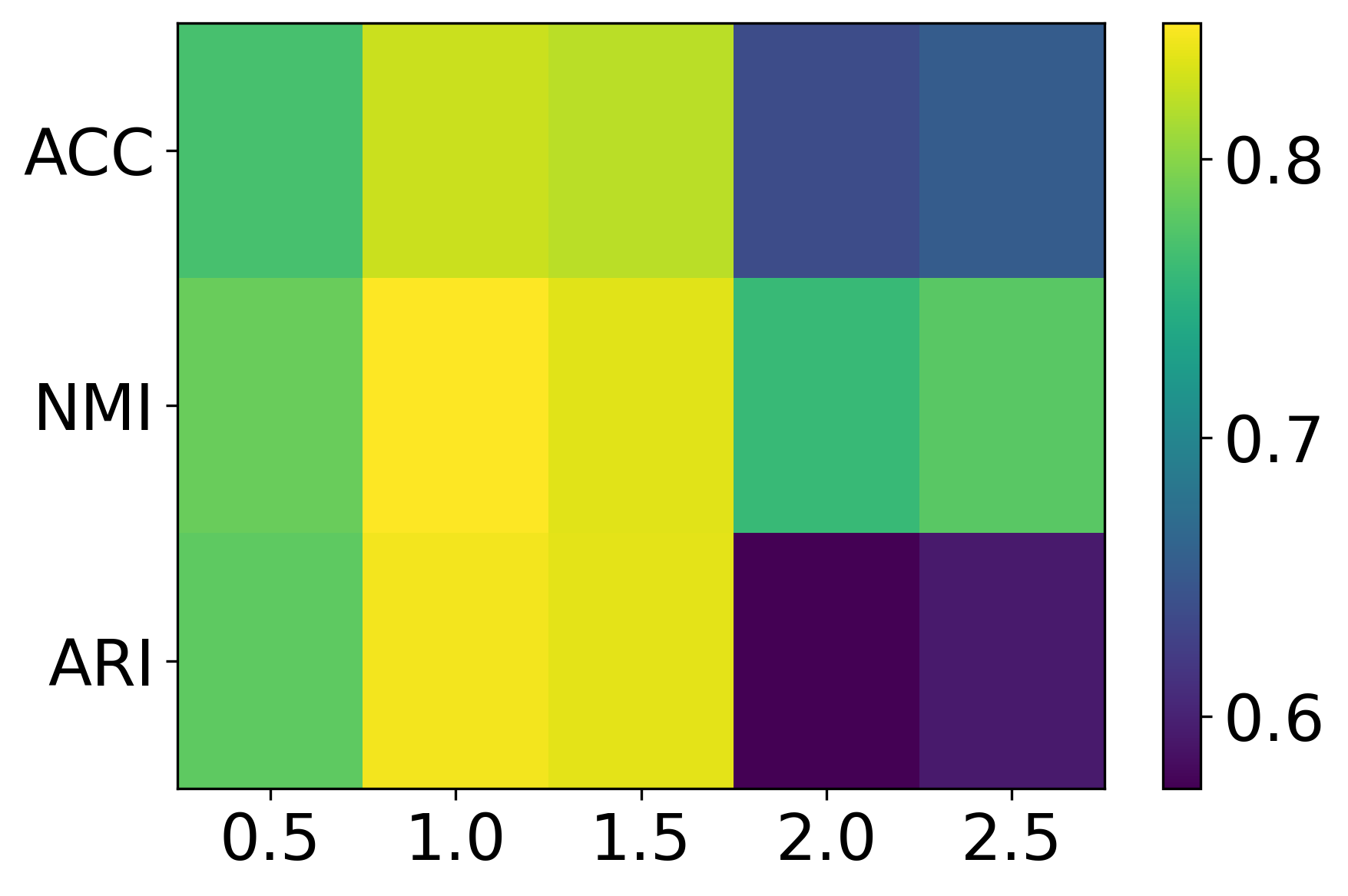}
        \caption{$\beta$}
        \label{fig:sens_beta}
    \end{subfigure}
    \hfill
    \begin{subfigure}[b]{0.49\linewidth}
        \centering
        \includegraphics[width=\linewidth]{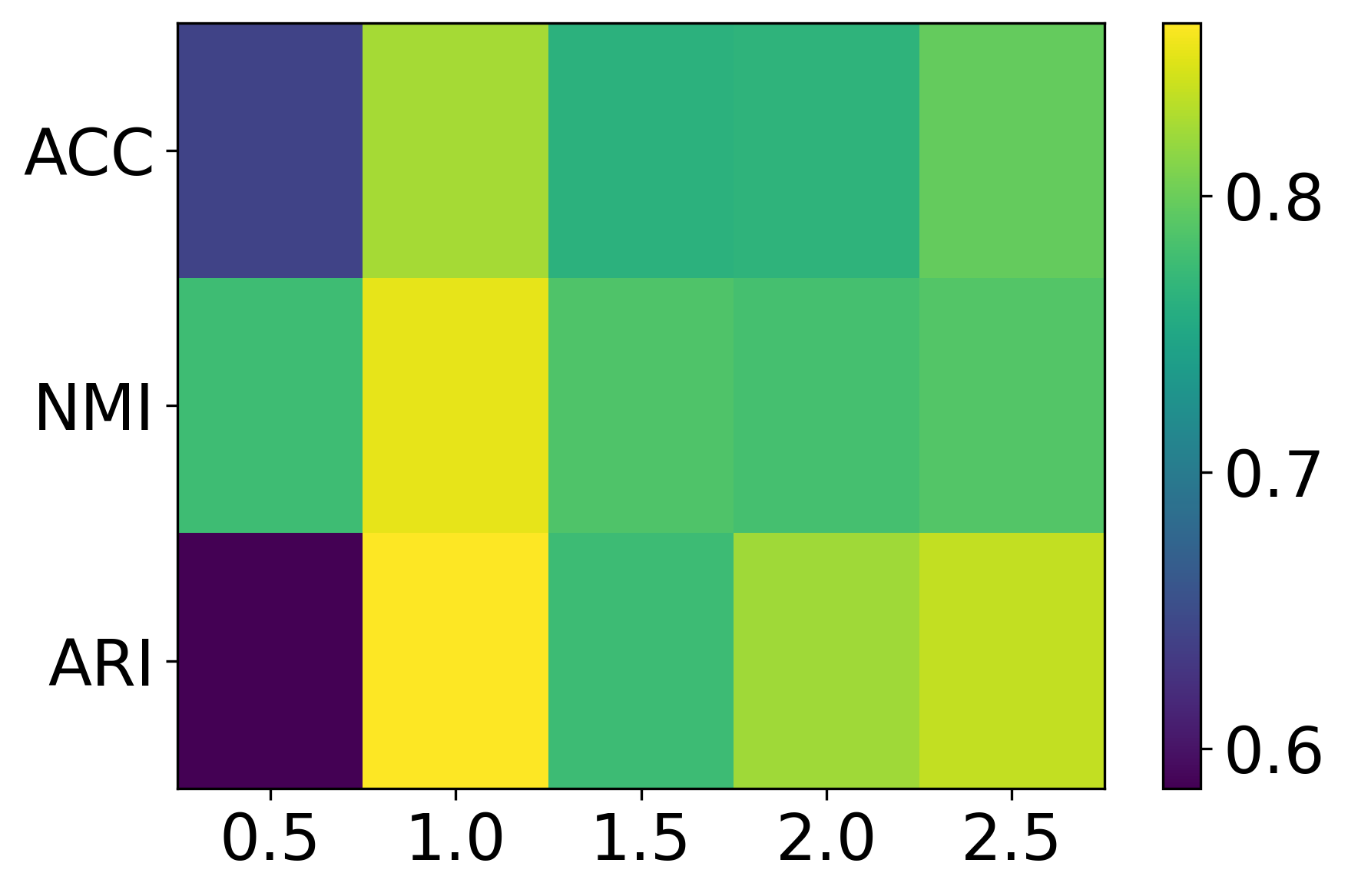}
        \caption{$\gamma$}
        \label{fig:sens_gamma}
    \end{subfigure}
    
    \caption{Impact of four parameters on clustering performance.}
    \label{fig:sensitivity_4params}
\end{figure}

To assess the robustness of \method{} against hyperparameter variations, we conduct a comprehensive sensitivity analysis on the Human Liver cells dataset. We specifically evaluate four key hyperparameters: the graph diffusion coefficient $\eta$ and three trade-off weights for the loss terms, denoted as $\alpha$ for the correlation loss, $\beta$ for the reconstruction loss, and $\gamma$ for the ZINB loss. As illustrated in Figure~\ref{fig:sensitivity_4params}, \method{} demonstrates remarkable stability across a broad range of parameter settings. Notably, the correlation weight $\alpha$  exhibit minimal impact on clustering accuracy, maintaining high performance even as values span multiple orders of magnitude. For the reconstruction and clustering constraints, optimal performance is consistently achieved when $\beta$ and $\gamma$ are set around 1.0. While extreme values can lead to performance degradation by over-emphasizing reconstruction at the expense of clustering structure, the model remains robust within a reasonable range. Regarding the graph diffusion coefficient $\eta$, a value between 0.2 and 0.3 yields the best results, effectively balancing local neighborhood information with global structural context. These results confirm that \method{} is not overly sensitive to specific hyperparameter tuning, making it practical for diverse scRNA-seq applications. We further report the sensitivity with respect to $K$ and $L$ measured by ARI. As shown in Figure~\ref{fig:sensitivity_ari}, the trend is consistent with that of ACC and NMI, confirming the robustness of \method{}. This consistency across ACC, NMI, and ARI confirms the strong performance of our methods across all benchmarks.


\section{Analysis of Runtime}
\label{sec:runtime}
\begin{figure}
    \centering
    \begin{subfigure}[b]{0.9\linewidth}
        \centering
        \includegraphics[width=\linewidth]{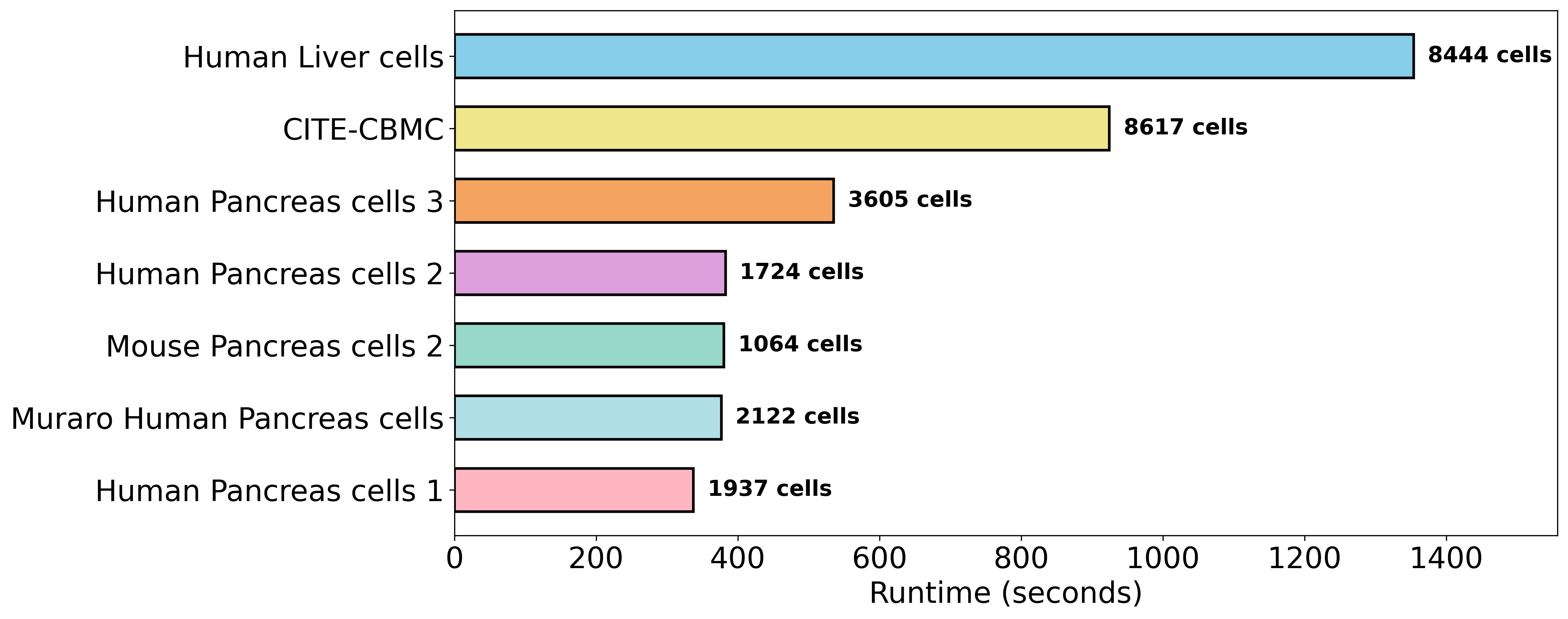} 
        \caption{Scalability analysis: Runtime vs. Number of Cells}
        \label{fig:runtime_scale}
    \end{subfigure}
    
    \vspace{0.5cm} 
    
    \begin{subfigure}[b]{0.9\linewidth}
        \centering
        \includegraphics[width=\linewidth]{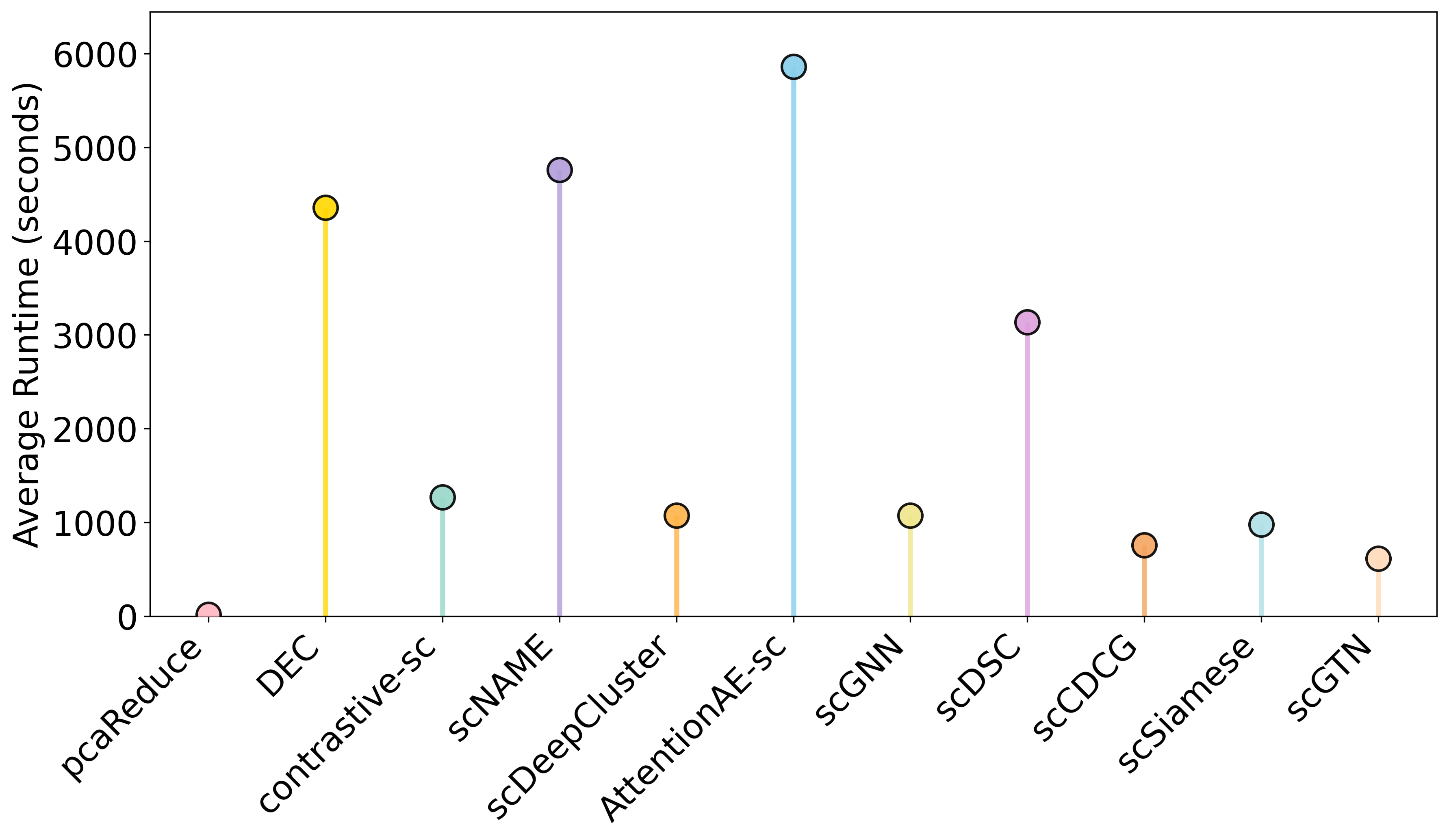}
        \caption{Runtime comparison with baseline methods}
        \label{fig:runtime_comp}
    \end{subfigure}
    
    \caption{(a) Scalability analysis of training time with respect to the number of cells. (b) Runtime comparison with baseline methods.}
    \label{fig:runtime}
\end{figure}

To evaluate the computational efficiency and scalability of \method{}, we conduct a comprehensive runtime analysis on datasets of varying scales. Regarding scalability with respect to data size, Figure~\ref{fig:runtime}(a) records the training time of \method{} on seven datasets. The results exhibit a near-linear relationship between runtime and the number of cells, demonstrating that our framework scales efficiently to large-scale scRNA-seq datasets. This efficiency is attributed to the graph construction and the optimized implementation of the Siamese Graph Transformer, which avoids the quadratic complexity typically associated with attention mechanisms. We further compare the runtime of \method{} against ten representative and competitive baselines. As depicted in Figure~\ref{fig:runtime}(b), even though \method{} explicitly incorporates complex structural priors, specifically shortest-path and position embeddings, its computational cost remains comparable to or even lower than other GNNs-based methods like scGNN. This indicates that our proposed \method{} achieves a superior trade-off between clustering performance and computational efficiency, ensuring practical feasibility for real-world applications.
\begin{table}[t]
\centering
\small
\begin{tabular}{cccc|ccc}
\toprule
\boldmath$\mathcal{L}_{\text{clu}}$ & \boldmath$\mathcal{L}_{\text{cor}}$ & \boldmath$\mathcal{L}_{\text{ZINB}}$ & \boldmath$\mathcal{L}_{\text{rec}}$ & \textbf{ACC} & \textbf{NMI} & \textbf{ARI} \\
\midrule
$\checkmark$ & $\checkmark$ & $\checkmark$ & $\times$      & 84.57 & 78.93 & 76.08 \\
$\checkmark$ & $\checkmark$ & $\times$      & $\checkmark$ & 84.18 & 78.36 & 75.62 \\
$\checkmark$ & $\times$      & $\checkmark$ & $\checkmark$ & 83.47 & 77.58 & 74.36 \\
$\times$      & $\checkmark$ & $\checkmark$ & $\checkmark$ & 81.24 & 74.79 & 70.68 \\
\midrule
$\checkmark$ & $\checkmark$ & $\times$      & $\times$      & 72.36 & 66.07 & 62.28 \\
$\checkmark$ & $\times$      & $\checkmark$ & $\times$      & 71.64 & 64.98 & 61.07 \\ 
$\checkmark$ & $\times$      & $\times$      & $\checkmark$ & 68.49 & 61.95 & 58.42 \\
$\times$      & $\checkmark$ & $\checkmark$ & $\times$      & 56.27 & 48.14 & 43.46 \\ 
$\times$      & $\checkmark$ & $\times$      & $\checkmark$ & 54.83 & 46.17 & 41.72 \\ 
$\times$      & $\times$      & $\checkmark$ & $\checkmark$ & 51.76 & 43.18 & 38.57 \\ 
\midrule
$\checkmark$ & $\checkmark$ & $\checkmark$ & $\checkmark$ & \textbf{96.02} & \textbf{89.15} & \textbf{93.10} \\
\bottomrule
\end{tabular}
\caption{Ablation experiment results of major loss terms}
\label{tab:main_ablation_more}
\end{table}

\section{Extended Ablation Study on Loss Terms}
\label{sec:extended_ablation}
We extend the ablation study to include scenarios where pairs of loss components are removed simultaneously. As detailed in Table~\ref{tab:main_ablation_more}, removing dual components leads to a distinct performance decline compared to single-component ablation. While single omissions maintain ACC above 80\%, dual omissions drop to the 50\%--72\% range, indicating strong interdependency among modules. Specifically, the results highlight the dominance of the optimal transport clustering loss $\mathcal{L}_{\text{clu}}$. Configurations retaining $\mathcal{L}_{\text{clu}}$ consistently outperform those excluding it  by approximately 15--20\%. Notably, relying solely on reconstruction objectives $\mathcal{L}_{\text{ZINB}}$ and $\mathcal{L}_{\text{rec}}$ yields the lowest performance, demonstrating that feature learning alone is insufficient for separating complex cell states without explicit clustering guidance. The full model's superior performance still confirms the necessity of jointly optimizing structural, feature, and distribution constraints.



\end{appendix}

\end{document}